\pdfoutput=1
\documentclass[11pt]{article}

\usepackage[]{acl}
\usepackage{times}
\usepackage{latexsym}
\usepackage{color}
\usepackage{algorithm,algpseudocode}
\usepackage{amsmath}
\usepackage{amsfonts}
\usepackage{xcolor,colortbl}
\usepackage{amssymb}
\usepackage{algpseudocode}
\usepackage{multirow}
\usepackage{tcolorbox}
\usepackage{ulem}
\usepackage{caption}
\usepackage{subcaption}

\definecolor{lightblue}{RGB}{225, 245, 254}
\definecolor{lightyellow}{RGB}{255, 253, 231}
\definecolor{lightgray}{RGB}{236, 239, 241}
\definecolor{lightgray}{rgb}{0.83, 0.83, 0.83}
\definecolor{lightbrown}{RGB}{239, 235, 233}
\definecolor{lightgreen}{RGB}{241, 248, 233}
\definecolor{lavenderpink}{rgb}{0.98, 0.68, 0.82}
\definecolor{lavenderindigo}{rgb}{0.58, 0.34, 0.92}
\definecolor{brilliantlavender}{rgb}{0.96, 0.73, 1.0}
\definecolor{columbiablue}{rgb}{0.61, 0.87, 1.0}
\newcommand\bluebox{\tcbox[on line,boxsep=3pt,left=1pt,right=1pt,top=1pt,bottom=1pt,boxrule=0.5pt,colback=lightblue,colframe=black]}
\newcommand\yellowbox{\tcbox[on line,boxsep=3pt,left=1pt,right=1pt,top=1pt,bottom=1pt,boxrule=0.5pt,colback=lightyellow,colframe=black]}
\newcommand\graybox{\tcbox[on line,boxsep=3pt,left=1pt,right=1pt,top=1pt,bottom=1pt,boxrule=0.5pt,colback=lightgray,colframe=black]}

\newcommand\greenbox{\tcbox[on line,boxsep=3pt,left=1pt,right=1pt,top=1pt,bottom=1pt,boxrule=0.5pt,colback=lightgreen,colframe=black]}

\algnewcommand\algorithmicforeach{\textbf{for}}
\algdef{S}[FOR]{For}[1]{\algorithmicforeach\ #1\ \algorithmicdo}

\usepackage{times}
\usepackage{latexsym}

\usepackage[T1]{fontenc}
\usepackage[utf8]{inputenc}

\usepackage{microtype}

\usepackage{inconsolata}

\usepackage{inconsolata}
\usepackage[utf8]{inputenc}
\usepackage[T1]{fontenc}
\usepackage{hyphenat}
\usepackage{xspace}
\usepackage{amsmath}
\usepackage{amsfonts}
\usepackage{hyperref}
\usepackage{url}
\usepackage{booktabs}
\usepackage{multirow}
\usepackage{makecell}
\usepackage{caption}
\usepackage{minibox}
\usepackage{bbm}
\usepackage{graphicx}
\usepackage{balance}
\usepackage{mathtools}
\usepackage{color}
\usepackage{marvosym}
\usepackage{ifthen}
\usepackage{textcomp}
\usepackage{enumitem}
\usepackage{verbatim}
\usepackage{algorithm}
\usepackage{numprint}
\usepackage{balance}

\usepackage{amsthm}
\theoremstyle{plain}

\newcommand{\chatoDisplayMode}[1]{#1}

\definecolor{MyRed}{rgb}{0.6,0.0,0.0} 
\definecolor{MyBlack}{rgb}{0.1,0.1,0.1} 
\newcommand{\inred}[1]{{\color{MyRed}\sf\textbf{\textsc{#1}}}}
\newcommand{\frameit}[2]{
  \begin{center}
  {\color{MyRed}
  \framebox[.9\columnwidth][l]{
    \begin{minipage}{.85\columnwidth}
    \inred{#1}: {\sf\color{MyBlack}#2}
    \end{minipage}
  }\\
  }
  \end{center}
}

\newcommand{\note}[2][]{\chatoDisplayMode{\def\@tmpsig{#1}\frameit{{\Pointinghand} Note}{#2\ifx \@tmpsig \@empty \else \mbox{ --\em #1}\fi}}}
\newcommand{\todo}[2][]{\chatoDisplayMode{\def\@tmpsig{#1}\frameit{{\Writinghand} To-do}{#2\ifx \@tmpsig \@empty \else \mbox{ --\em #1}\fi}}}

\newcommand{\abbrevStyle}[1]{#1}
\newcommand{\etc}{\abbrevStyle{etc.}\xspace}

\newcommand{\Tabref}[1]{Table~\ref{#1}}
\newcommand{\Figref}[1]{Fig.~\ref{#1}}

\newcommand{\Algref}[1]{Alg.~\ref{#1}}

\newcommand{\textcite}[1]{\citeauthor{#1} \shortcite{#1}}

\newcommand{\hide}[1]{}

\makeatletter
\newcommand{\iffont}[2]{\ifthenelse{\equal{\f@family}{#1}}{#2}{}}
\makeatother

  \usepackage{mathptmx}

  \DeclareSymbolFont{greek}{OML}{cmm}{m}{n}
  \DeclareMathSymbol{\alpha}{\mathalpha}{greek}{"0B}
  \DeclareMathSymbol{\beta}{\mathalpha}{greek}{"0C}
  \DeclareMathSymbol{\gamma}{\mathalpha}{greek}{"0D}
  \DeclareMathSymbol{\delta}{\mathalpha}{greek}{"0E}
  \DeclareMathSymbol{\epsilon}{\mathalpha}{greek}{"0F}
  \DeclareMathSymbol{\zeta}{\mathalpha}{greek}{"10}
  \DeclareMathSymbol{\eta}{\mathalpha}{greek}{"11}
  \DeclareMathSymbol{\theta}{\mathalpha}{greek}{"12}
  \DeclareMathSymbol{\iota}{\mathalpha}{greek}{"13}
  \DeclareMathSymbol{\kappa}{\mathalpha}{greek}{"14}
  \DeclareMathSymbol{\lambda}{\mathalpha}{greek}{"15}
  \DeclareMathSymbol{\mu}{\mathalpha}{greek}{"16}
  \DeclareMathSymbol{\nu}{\mathalpha}{greek}{"17}
  \DeclareMathSymbol{\xi}{\mathalpha}{greek}{"18}
  \DeclareMathSymbol{\pi}{\mathalpha}{greek}{"19}
  \DeclareMathSymbol{\rho}{\mathalpha}{greek}{"1A}
  \DeclareMathSymbol{\sigma}{\mathalpha}{greek}{"1B}
  \DeclareMathSymbol{\tau}{\mathalpha}{greek}{"1C}
  \DeclareMathSymbol{\upsilon}{\mathalpha}{greek}{"1D}
  \DeclareMathSymbol{\phi}{\mathalpha}{greek}{"1E}
  \DeclareMathSymbol{\chi}{\mathalpha}{greek}{"1F}
  \DeclareMathSymbol{\psi}{\mathalpha}{greek}{"20}
  \DeclareMathSymbol{\omega}{\mathalpha}{greek}{"21}
  \DeclareMathSymbol{\varepsilon}{\mathalpha}{greek}{"22}
  \DeclareMathSymbol{\vartheta}{\mathalpha}{greek}{"23}
  \DeclareMathSymbol{\varpi}{\mathalpha}{greek}{"24}
  \DeclareMathSymbol{\varrho}{\mathalpha}{greek}{"25}
  \DeclareMathSymbol{\varsigma}{\mathalpha}{greek}{"26}
  \DeclareMathSymbol{\varphi}{\mathalpha}{greek}{"27}
  \DeclareSymbolFont{otone}{OT1}{cmr}{m}{n}
  \DeclareMathSymbol{\Gamma}{\mathalpha}{otone}{0}
  \DeclareMathSymbol{\Delta}{\mathalpha}{otone}{1}
  \DeclareMathSymbol{\Theta}{\mathalpha}{otone}{2}
  \DeclareMathSymbol{\Lambda}{\mathalpha}{otone}{3}
  \DeclareMathSymbol{\Xi}{\mathalpha}{otone}{4}
  \DeclareMathSymbol{\Pi}{\mathalpha}{otone}{5}
  \DeclareMathSymbol{\Sigma}{\mathalpha}{otone}{6}
  \DeclareMathSymbol{\Upsilon}{\mathalpha}{otone}{7}
  \DeclareMathSymbol{\Phi}{\mathalpha}{otone}{8}
  \DeclareMathSymbol{\Psi}{\mathalpha}{otone}{9}
  \DeclareMathSymbol{\Omega}{\mathalpha}{otone}{10}
  \DeclareSymbolFont{syms}{OML}{cmm}{m}{it}
  \DeclareMathSymbol{\partial}{\mathord}{syms}{"40}
  \DeclareMathAlphabet{\mathbold}{OML}{cmm}{b}{it}
  \DeclareSymbolFont{largesymbols}{OMX}{cmex}{m}{n}

\usepackage{amsmath}
\usepackage{amssymb}

\newcommand{\ourmodel}{REFINER\xspace}
\definecolor{MyBlue}{rgb}{0.25,0.5,0.75}
\definecolor{blush}{rgb}{0.87, 0.36, 0.51}

\colorlet{NextBlue}{MyBlue!20}
\colorlet{NextBlue}{MyBlue!20}
\colorlet{SecondBlue}{MyBlue!40}
\colorlet{Nextblush}{blush!30}
\newcommand{\model}{{REFINER}\xspace}

\title{\ourmodel: Reasoning Feedback on Intermediate Representations}

\author{Debjit Paul$^\spadesuit$, 
  Mete Ismayilzada$^\spadesuit$,  
  Maxime Peyrard$^\diamondsuit$$^*$, 
  Beatriz Borges$^\spadesuit$, \\ 
  \textbf{Antoine Bosselut$^\spadesuit$}, 
  \textbf{Robert West$^\spadesuit$},
  \textbf{Boi Faltings$^\spadesuit$} \\
  $^\spadesuit$EPFL \\ 
  $^\diamondsuit$Université Grenoble Alpes, CNRS, Grenoble INP, LIG \\
  \texttt{\{firstname.lastname\}@epfl.ch}}
\begin{document}
\maketitle
\begin{abstract}

Language models (LMs) have recently shown remarkable performance on reasoning tasks by explicitly generating intermediate inferences, e.g., chain-of-thought prompting.  However, these intermediate inference steps may be inappropriate deductions from the initial context and lead to incorrect final predictions. Here we introduce \ourmodel, a framework for finetuning LMs to explicitly generate intermediate reasoning steps while interacting with a critic model that provides automated feedback on the reasoning. Specifically, the critic provides structured feedback that the reasoning LM uses to iteratively improve its intermediate arguments. Empirical evaluations of \ourmodel{ } on three diverse reasoning tasks show significant improvements over baseline LMs of comparable scale. Furthermore, when using GPT-3.5 or ChatGPT as the reasoner, the trained critic significantly improves reasoning without finetuning the reasoner. Finally, our critic model is trained without expensive human-in-the-loop data but can be substituted with humans at inference time. 
 
\end{abstract}

\section{Introduction} \label{sec:introduction}

Large language models (LLMs) have made significant strides in natural language processing (NLP) tasks \citep{NEURIPS2020_1457c0d6}. Recent work has shown that explicitly generating intermediate steps during reasoning tasks significantly improves a model's performance and interpretability \citep{shwartz-etal-2020-unsupervised, paul-frank-2021-coins, marasovic-etal-2022-shot, lampinen2022tell, wei2022chain}. Producing such intermediate representations provides insight into the model's predictions and allows
humans to inspect the model's reasoning process. However, these intermediate representations\footnote{In a reasoning task, the intermediate representations can be viewed as inference rules, explanations or reasoning steps. \\ * Work done at EPFL} can be unreliable \citep{ye2022the} and  result in poor performance on downstream reasoning tasks. Most importantly, it is unclear how to meaningfully refine the intermediate representations to further improve the final performance.

\begin{figure}[t]
    \centering   
    \includegraphics[scale=0.6,height=6.5cm, width=0.325\paperwidth]{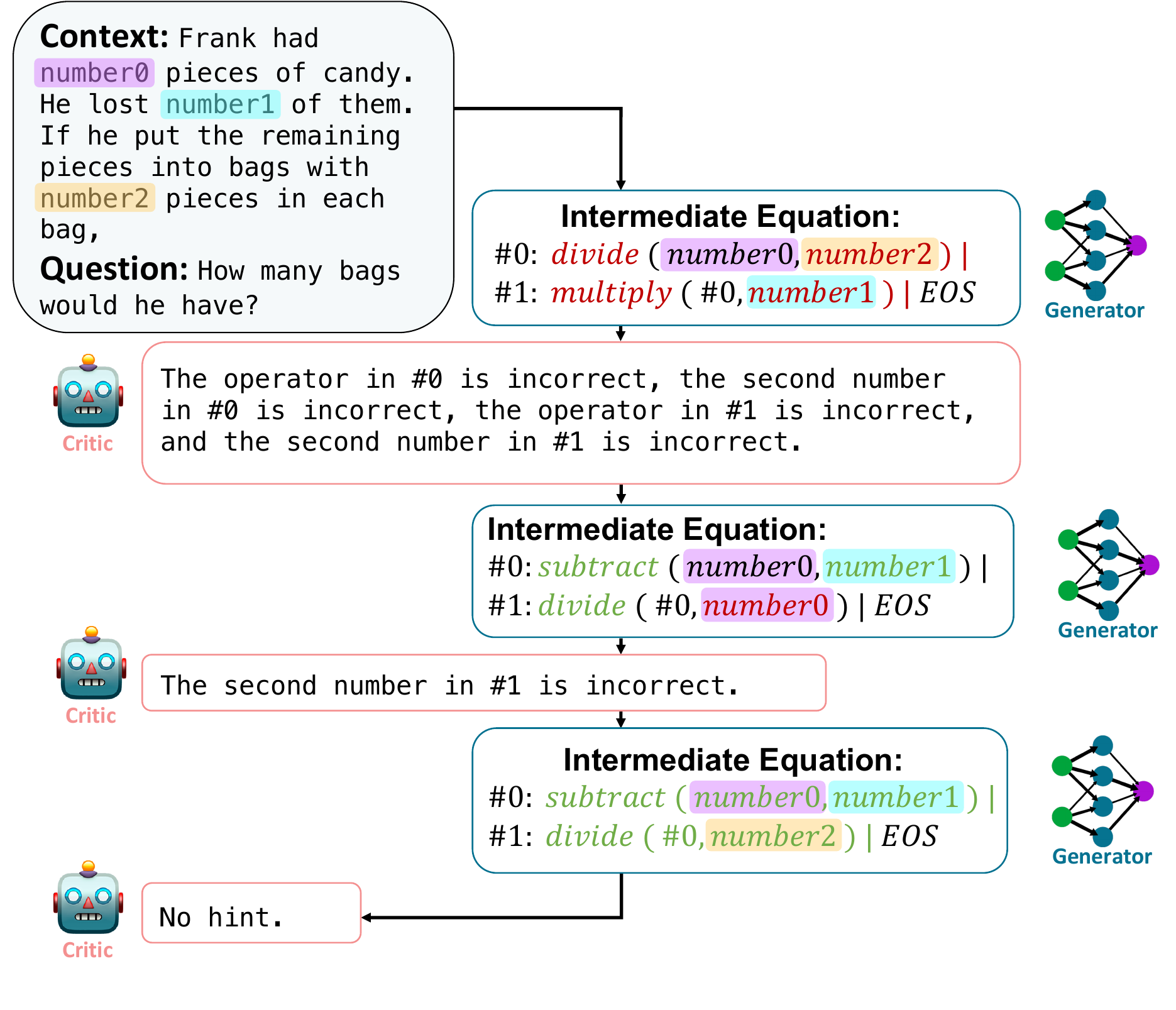}
    \caption{\textbf{\ourmodel example.} The critic model provides the generator model with feedback on its reasoning errors after evaluating the generated intermediate steps. The feedback, alongside the original question and previous intermediate equation, are fed back to the generator model.}
    \label{fig:model_example}
    \vspace{-2em}
\end{figure}

The standard practice for correcting reasoning errors is to annotate new data and either retrain or finetune the model \citep{feng-etal-2021-survey, hedderich-etal-2021-survey}. However, fixing such errors by finetuning with more data is not only data- and resource-intensive but can also be insufficient to generalize well in complex reasoning tasks \citep{Ward2022ArgumentativeRL}. Other works have explored improving models using feedback by providing a scalar reward  \citep{ziegler2019finetuning, martin-etal-2022-learning} or directly revealing the correct missing answer \citep{mehta-goldwasser-2019-improving, elgohary-etal-2021-nl, tandon-etal-2022-learning}. However, in natural language reasoning tasks, defining a reward that captures different fine-grained reasoning error types (\textit{e.g.,} semantic consistency, logical, \etc) remains an open challenge \citep{anonymous2023roscoe}. Additionally, such a reward provides a relatively sparse training signal.

In this work, we instead provide fine-grained and structured feedback on reasoning errors. We present \ourmodel, a novel interaction-based framework that allows a generator LM to iteratively use fine-grained feedback and refine its reasoning. The interaction happens between two models: a \textit{generator}, which learns to solve the task by first generating the intermediate reasoning steps, and a \textit{critic}, which provides structured feedback to the generator about errors in the intermediate steps. 
 
To provide fine-grained feedback about reasoning errors, we develop a scheme to independently train the critic model on automatically constructed feedback data. More specifically, we create pairs of incorrect intermediate representations and structured\footnote{Note that we transform the structured feedback into semi-structured textual feedback using templates.} feedback on their fine-grained reasoning errors. Then, we use this data to train the critic to provide fine-grained feedback on erroneous intermediate reasoning steps.
Finally, the critic interacts with the generator LM, offering feedback both during the training of the generator and during inference. 

Figure \ref{fig:model_example} illustrates an example of our \ourmodel framework where, given a math word problem, the generator generates an equation as an intermediate representation. The critic identifies the errors in the equation and provides semi-structured textual feedback (\textit{e.g.,} "\texttt{the operator in} $\#0$ \texttt{is incorrect}") to the generator. By interacting with the critic, \ourmodel enables the generator to reason over the semi-structured feedback and \textit{refine} its generation.

\textbf{Contributions.}
(i) We propose \ourmodel, a framework that refines LMs reasoning capabilities through feedback.  
% To the best of our knowledge, 
Our work 
% is the first to 
investigates how interacting with fine-grained reasoning feedback on intermediate reasoning steps impacts the performance of LMs on reasoning tasks. We evaluate \ourmodel on three natural language reasoning tasks: math word problems, synthetic natural language reasoning, and moral action generation. \ourmodel demonstrates significant performance gains across different LM architectures with different scales. Across different reasoning tasks, \ourmodel outperforms comparably-sized strong fine-tuned LM baselines (by +13.1, +3.2, +15 pts., respectively). 
%improve the performance of relatively small LMs (\textit{e.g.,} T5-base) over strong baselines by ... 
(ii) We empirically demonstrate that for math word problems and synthetic natural language reasoning, our trained critic models alone are beneficial for improving intermediate representations as they help GPT-$3.5$ significantly increase its performance in a few-shot setting (by +3.5, +6.8 pts., respectively). %Generating intermediate representations that are critiqued results in models with stronger reasoning capabilities that are interpretable with faithful intermediate reasoning steps. Furthermore, the trained critic can be substituted at test time by a human critic, yielding effective model-human interactions. 
We also demonstrate that providing structured feedback on fine-grained errors can benefit more than scalar value feedback for moral action generation and math word problem tasks. Our critic model acts as a `reasoning refinement tool' for LLMs. (iii) We show that REFINER can substantially outperform other refinement methods that use feedback from large LMs, such as self-refine. (iv) Our analyses illustrate that (a) improving the intermediate representation generation improves the performance on the reasoning tasks, and (b) training a generator with an imperfect (noisy) critic is still beneficial. Our code is made publicly available \footnote{\url{https://github.com/debjitpaul/refiner}}.
\section{Related Work}

\textbf{Intermediate Representations.} 
While state-of-the-art LMs achieve incredible performances in a wide range of tasks, they have difficulty with many reasoning tasks \cite{wang2022consistency}, especially ones with multiple constraints or sub-problems
% , similar keywords to more common tasks, 
or requiring specialized knowledge \cite{austin2021programsynth} -- such as mathematical problem solving \cite{ling-etal-2017-program, andor-etal-2019-giving, ran-etal-2019-numnet, geva-etal-2020-injecting, piekos-etal-2021-measuring, cobbe2021verifiers,kim-etal-2022-ept}.

For these tasks, both intermediate representations and rationales have been shown to be beneficial in learning mathematical skills \cite{piekos-etal-2021-measuring},  intermediate program execution computations \cite{nye2021scratchpads}, or general reasoning outputs \cite{wei2022chain, golovneva2022roscoe}. 

Our work builds upon the observation that generating intermediate steps are valuable but distinguishes itself in several key aspects. Firstly, instead of prompting a large model, we finetune smaller models to learn to generate intermediate steps. Secondly, our framework can accommodate tasks that do not necessarily have unique closed-form correct answer, such as the \textit{Moral Norm} task (see \S\ref{sec:model}). Finally, our framework is trained with a critic providing feedback, improving the model's reasoning process and teaching it how to leverage feedback.

\textbf{Natural Language Feedback.}
Recent work has explored giving models richer and more complex feedback through the use of natural language \cite{ziegler2019finetuning, nguyen2021interactive, scheurer2022nlfeedback}, used for aligning LLMs' output with users' preferences %(\eg, aiming to prevent misinformation or factually incorrect information) 
\cite{christiano2017deepRL, ziegler2019finetuning, saunders2022selfcritiquing, scheurer2022nlfeedback, bai2022constitutional}, or to directly improve the model's performance in its current task \cite{weston2016dialog, rupprecht2018guide, elgohary-etal-2020-speak, austin2021programsynth, madaan2023selfrefine}.
This training depends on human-created feedback, generated in large quantities \cite{bai2022constitutional}, which takes up considerable resources. Though an external feedback provider can guide models to correct answers and reasoning \cite{austin2021programsynth}, demonstrably better than they can themselves \cite{saunders2022selfcritiquing}, feedback has rarely been used in this way -- and automated critics for reasoning tasks have proved to be difficult \cite{scheurer2022nlfeedback, wang2022consistency, huang2022selfimprove}. 

Recently, \citet{welleck2022generating} introduced a secondary model, the corrector, which improves the initial proposition of a generation model, by learning the kind of mistakes made by the generator and how to fix them. 
In this work, we also use a secondary model, a critic, but apply it quite differently as we integrate it into an interaction loop with the generator model during training. We further differ from previous works as we provide feedback at the intermediate reasoning steps of the model and not at the final output. The feedback is thus closer to the source of mistakes and guides the model's reasoning toward the correct answer. Additionally, intermediate steps are often structured% (\eg in math reasoning, they could be equations)
, allowing the critic to provide precise feedback.

\section{\ourmodel}
\label{sec:model}

\textbf{Problem Formulation.} In this paper, we view \textit{natural language reasoning} (NLR) as an autoregressive generation task where, given input context $x$, a model needs to generate $y$, such that $y$ satisfies the constraints of the task. 
Usually, to generate correct or plausible $y$, the model needs to make the correct inference $z$ as intermediate steps.\footnote{ We use ``inference steps/representations'' and ``hypothesis'' interchangeably.} 
We decompose NLR tasks as follows: 
% \begin{equation}
    $p(y|x) = p(y|x,z) p(z|x)$.
% \end{equation}
In practice, one can compute each conditional using an LM that includes its conditioning variables as a part of its input.

Before continuing with the model description, %we illustrate, in  Figure~\ref{fig:error_overview}, 
we describe three NLR tasks where we conduct our study and their respective intermediate representation $z$. We deliberately chose these three tasks since they broadly cover two types of reasoning: (i) logical reasoning and (ii) normative reasoning. They are exemplified in Appx ~\Figref{fig:error_overview} and detailed below. 

\textbf{Math word problem (MWP)}, where given a word problem $x$ consisting of a context and question, the goal is to map $x$ to a valid mathematical expression $z$ (the intermediate representation) and then to a solution $y$. This task requires the model to perform deduction using mathematical reasoning. \\
\textbf{Synthetic natural language reasoning (sNLR)}, where given a reasoning scenario $x$ consisting of $5$ synthetic rules and a fact, the model needs to deduce a conclusion $y$. 
This task requires the model to perform deductive reasoning and generate intermediate steps $z$ and the conclusion $y$ using closed-world rules and facts.\\ 
\textbf{Moral norm and action generation for moral stories (MS)}, where given a context $x$ consisting of a \textit{situation}, an \textit{intention}, and an \textit{immoral action}, the model needs to generate the moral norm  ${z}$ and the moral action ${y}$. Moral actions are encouraged by the moral norm. This task requires the model to perform abductive reasoning to generate moral norms and deductive reasoning for moral action.

We propose to solve these tasks by forcing the model to generate intermediate hypotheses ($z$) and improving them via structured feedback. We introduce an interactive framework, \ourmodel, made of two separate models: 
(a) a \textsc{critic} model (\S \ref{sec_3.1:critic_model}) trained to provide structured feedback on intermediate reasoning steps and (b) a \textsc{generator} model trained to solve the reasoning task by first generating intermediate reasoning steps (\S \ref{sec:3.2_generator}). 
The core idea of \ourmodel is to exploit the interaction between the generator model and the critic model, where the generator's intermediate reasoning steps are improved via structured feedback from the critic. %An overview of the framework is depicted in \Figref{fig:refiner_model}.

%An overview of the two tasks tackled in this paper, with examples of both valid and invalid intermediate reasoning steps, as well as their corresponding fine-grained error types.
\ourmodel presents several important properties. First, the generator is trained to incorporate and leverage feedback, which helps it converge towards better reasoning during training and makes it capable of integrating feedback at test time, whether from a trained critic or a human (see \S\ref{sec:result}).
Second, the trained critic can be useful on its own; we demonstrate that a generalist LLM like GPT-$3.5$ can significantly benefit from interacting with our trained critic on the reasoning tasks we consider (see \S\ref{sec:result}).
Finally, having two separate models allows us to easily measure the benefits of feedback during training and/or during inference (see \S\ref{sec:analysis}). 
    
\subsection{CRITIC Model}\label{sec_3.1:critic_model}

The role of the critic is to provide feedback on the intermediate hypotheses produced by the generator model. One way to evaluate the quality of the hypothesis and produce feedback on the hypothesis $z$, would be to compare it against a gold hypothesis $z^*$.  Previous works employed 
automatic metrics like BLEU, ROUGE, etc., as value functions \citep{wu-etal-2018-study, Ramamurthy2022IsRL}. However, these scalar value functions are not suitable for natural language reasoning tasks because (i) it is unclear how to define a scalar value function that can encapsulate fine-grained reasoning errors \citep{anonymous2023roscoe} and (ii) during inference, these functions require access to the gold hypothesis (which is unavailable in practice). Therefore, we train a critic model and endow it with the ability to evaluate the hypothesis in a fine-grained manner and provide structured feedback.

\textbf{Feedback Data Generation.} To train the critic, we have to create example pairs of implausible hypotheses and their corresponding feedback with fine-grained reasoning errors. Inspired by \newcite{anonymous2023roscoe} and \newcite{NEURIPS2020_e992111e}, we first define fine-grained reasoning error types for each reasoning task (see Table \ref{tab:define_error_feedbacks}). 
For MWP, an equation can be incorrect due to: (i) the operands or operators in the equations being incorrect and/or (ii) one or more operators missing. For sNLR, an inference rule can be incorrect because it is (i) logically invalid and/or (ii) missing reasoning rules (\textit{failing to connect the correct facts with correct rules or missing implicit knowledge}). For MS, a moral norm can be incorrect due to (i) contradiction and/or (ii) semantic misalignment. 

\begin{table}[t!]
\centering
\scalebox{0.65}{
\begin{tabular}{@{}l@{~~}l@{~~}l@{~~~}}
\toprule
{\bf Tasks} & {\bf Error Types} & {\bf Feedbacks}\\\midrule
%\rowcolor{gray!25}
%\textbf{Full training data} &  &\\

& {Incorrect Numbers} & The \texttt{position}  number in \\ 
&   & \texttt{equation-number} is incorrect. \\

\textbf{MWP} &{Incorrect Operators} & The operator in \\
&   & \texttt{equation-number} is incorrect. \\ 
 
&Missing Operators & An operator is missing.\\
\midrule
& {Logically Invalid} & The \texttt{X operator}  makes \texttt{inference}\\ 
&&\texttt{rule number} invalid. \\
\textbf{sNLR} & {Missing Link} & Missing link between the fact the rules. \\ 
& {Missing Implicit } & The implicit knowledge is \\ 
&  Knowledge Step & missing. \\
\midrule
& {Contradiction} & {Contradiction} \\
\textbf{MS} & Semantic Misalignment & Semantically misaligned:  
`` text snippet''\\
\bottomrule
\end{tabular}}
\caption{An overview of the Error Types and Feedbacks for each reasoning tasks.
}
\vspace{-1.5em}
\label{tab:define_error_feedbacks}
\end{table}

Based on these error types, we propose two strategies to create the feedback data: (i) \textbf{Rule-based perturbation} strategy: we perturb the plausible hypotheses ($z$) in the training data and collect a pool of data $D$ ($x$: input, $z$: plausible hypothesis, $z'$: implausible hypothesis). We perturb by omitting, replacing or adding some tokens or some rules from the plausible hypothesis to create an implausible hypothesis automatically (details in Appendix \ref{sec:perturbation_gen}). %We generated $1.5$M, $100$K, and $50$K implausible hypotheses for MWP, sNLR and MS tasks, respectively. 
% in Fig.\ref{fig:error_overview}, for sNLR we omit a few inference steps from the correct hypothesis "\texttt{\#0: viridian is green, \#1: rose is green}" and create an incorrect (incomplete) hypothesis (see \Figref{fig:error_overview}).
(ii) \textbf{Synthetic Generation} strategy: we prompted OpenAI's GPT-$3.5$ to generate implausible hypotheses based on the error types automatically. We used a few-shot setting where we varied the
instruction, the number of demonstrations, and the formatting of the demonstrations (details in Appendix \ref{sec:synthetic_gen}). %Since data generation with GPT-$3.5$ is expensive, we generated $30$K, $20$K, and $30$K implausible hypotheses for MWP, sNLR and MS tasks, respectively. 

Since our perturbations and automatic implausible hypotheses are based on logic and reasoning errors, we create structured feedback $f$ for every example ($x, z, z'$) by stating the error type that occurs in $z'$ but not in $z$ (see Table \ref{tab:define_error_feedbacks}). The basic structure of feedback $f$ for these tasks is $\langle$\textit{error type, position (optional), hint (optional)}$\rangle$, where position denotes the error position in the implausible hypothesis (see Table \ref{tab:define_error_feedbacks}). %For example, in the previous scenario, we create feedback ``\textit{Missing link between fact and rules}''. 
Despite the simplicity of the strategy we used for our tasks, this approach is easily generalisable to other reasoning tasks. 

\if False 
For MWP and sNLR problems, the underlying reasoning requires symbolic systems with closed-world rules. Hence, we consider a simple rule-based method to automatically generate the pairs of errors and their corresponding structured feedback by considering the error types and position of the errors (see \Figref{fig:error_overview} and Table \ref{tab:define_error_feedbacks}). 

In the moral norm generation task, we consider two kinds of fine-grained errors: \textit{logical contradiction} and \textit{semantic misalignment} (incoherent, uninformative). Moral norms are people’s subjective judgments about the character and actions mentioned in the context. Each moral norm is a combination of two components (implicit structure): a moral judgment \texttt{[You shouldn't]} and an action \texttt{[criticize your family's religion]}. 
Firstly, to create \textit{logical contradictions}, we use the concept of deontic logic from \citet{kiehne-emnlp-2022} and derive new norms contrary to those of Moral Stories. Hence, we replace the correct moral judgments in the plausible hypothesis with inverse judgments. For example, replacing \texttt{[You shouldn't]} from the plausible hypothesis to \texttt{[It's good]}, as depicted in \Figref{fig:error_overview}. To scale such inverse norms (\textit{implausible hypothesis}), we paraphrase them by substituting the adjectives with synonyms from WordNet. 
Secondly, to create \textit{semantic misalignments}, we must collect implausible hypotheses that are either misaligned with the plausible hypothesis or incomplete in nature. 
To create them, we replace the correct action (verb phrase) from the plausible hypothesis with random verb phrases selected from the context of the plausible hypothesis. 
\fi 
\begin{figure}[t]
  \centering
    \includegraphics[scale=1, width=0.35\paperwidth]{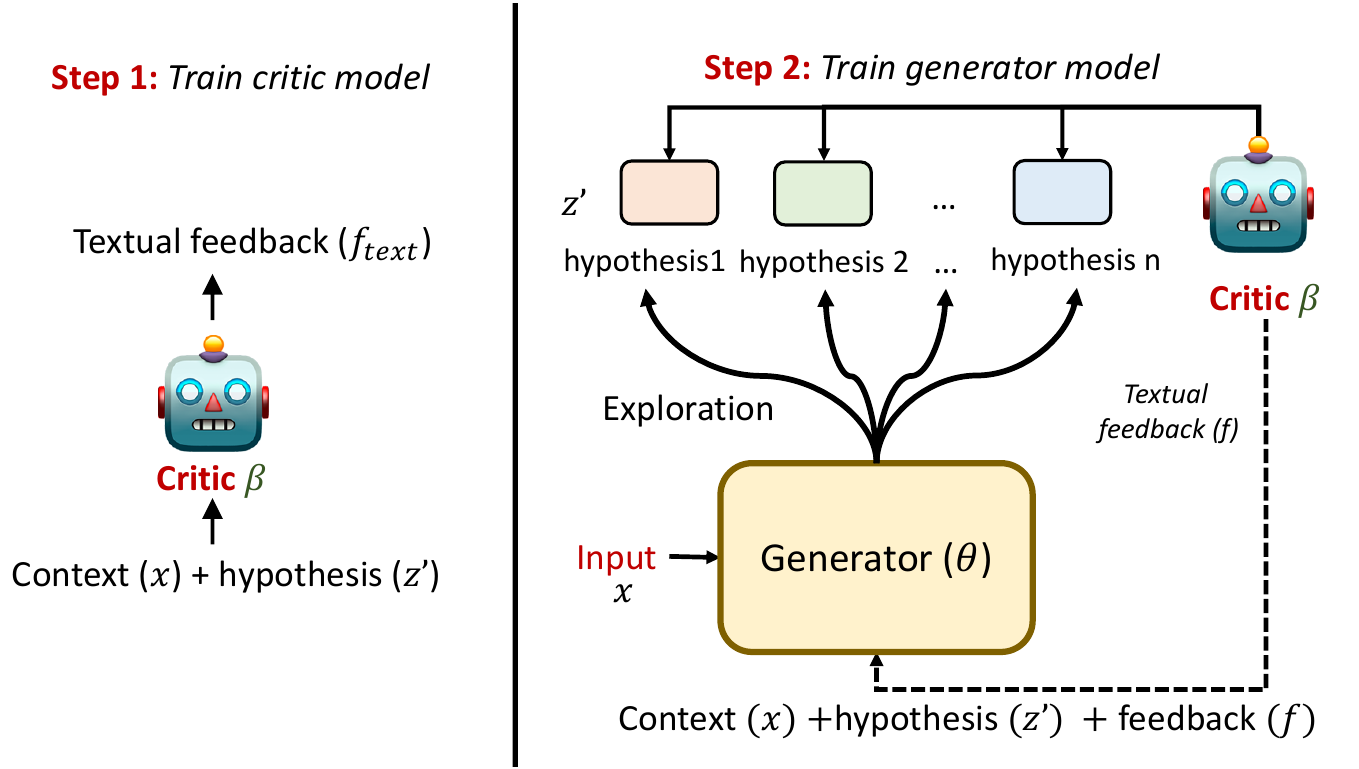}
    \caption{Overview of \ourmodel interaction loop. Left side: Training the critic model. Right side: In each iteration, the generator generates multiple hypotheses. The critic randomly selects one hypothesis and provides feedback based on reasoning errors.}
    \label{fig:refiner_model}
    \vspace{-1.5em}
\end{figure}
%Notice the \textbf{Missing Steps} error type, in the second task, actually encompasses two error types: reasoning misalignment, derived from not considering the \texttt{or} operation, and lack of implicit knowledge, where implicit knowledge is needed to match the existing rules. }
We also replace the correct judgment with random judgments to scale the number of implausible hypotheses per example. Finally, as feedback $f$, we provide $<$\textit{error type, hint}$>$. For non-monotonic reasoning tasks like norm and action generation, the critic should be able to provide hints that align the generator model's objective to the reasoning task. Hence, as a \textit{hint}, we provide verb phrases from the norms. 
Since the critic provides textual feedback to the generator, we convert the structured feedback into natural language feedback \footnote{Further details about feedback are provided in Appx.\ref{sec:feedback_gen}.}. Formally, we create a data pool $D = \{x, z, z', f\}$ to train a critic model.

\textbf{Training the critic model.} We train a supervised \textsc{critic} model ($\pi_{\beta}$) with the context ($x$) and (plausible or implausible) hypothesis ($z$ or $z'$) as input and the textual feedback as output.
We update the \textsc{critic} with the cross-entropy loss: 
%\begin{equation}\label{eq:critic}
$    L(\beta) = -\log p_{\beta}(f(u)|x, u)$
%\end{equation}
where $u \in z, z'$. The trained critic is only used during inference. 
The oracle critic is used while training the generator.
% model. 

\subsection{GENERATOR Model} \label{sec:3.2_generator}

This section presents a generator model that iteratively learns to interact with the \textsc{critic} model. 

\textbf{Warm-up.} Given a context $x$ the generator model ($\pi_{\theta}$) is trained to generate plausible hypotheses. The warm-up phase is critical to ensure that, when the critic comes in the loop, the generator does not produce random answers likely to be bad, given the size of the output space. As such, we use a small supervised dataset (10\% training data) to fine-tune the model on the NLR task of interest. After the warm-up phase, we use the additional feedback $f$ from the critic model and learn $\pi_{\theta}(z|x, z', f)$.

\textbf{Exploration.} At each iteration ($t$), the generator model generates multiple hypotheses ($z^k$) using nucleus sampling. The critic model randomly selects one hypothesis and provides feedback on that hypothesis. The exploration step aims at increasing the output variance such that the generator receives a wide range of feedback during training. 

\textbf{Learning.} We update the \textsc{generator} model using the following cross-entropy loss: 
%\begin{equation}
$    L(\theta) = -\sum_{t=1}^{T}\log p_{\theta}(z_t| x, z_t', f_t(z'))$
%\end{equation}  
where $T$ = total number of iterations. Since the feedback contains the error types and hints, which are (latent) fine-grained and logical, it should allow the model to learn and update its generation by addressing the reasoning errors mentioned in the feedback. 

%\begin{algorithm}\caption{Online Learning}
%\begin{algorithmic}[1]
%\For {$i \gets 1$ to $T$}                  
%    \State {$\hat{y}_{i} \gets \pi_{\theta}(y_i|q_i, c_i)$}
%    \State {Observe a reward $r_{i}$}
%    \State {Update the model parameters $\theta$ using the gradient:\\} 
%    {$r_{i} \nabla_{\theta}$ log $\pi_{\theta}(\hat{y}_i|q_i, c_i)$}
%\EndFor
%\end{algorithmic}
%\end{algorithm}

%\begin{algorithm}\caption{Train a Critique}
%\begin{algorithmic}[1]
%\For {$i \gets 1$ to $T$}                  
%    \State {$\hat{y}_{i} \gets \pi_{\theta}(y_i|q_i, c_i)$}
%    \State {Observe a reward $r_{i}$}
%    \State {Update the model parameters $\theta$ using the gradient:\\} 
%    {$r_{i} \nabla_{\theta}$ log $\pi_{\theta}(\hat{y}_i|q_i, c_i)$}
%\EndFor
%\end{algorithmic}
%\end{algorithm}

\textbf{Inference.} We use the trained critic 
% (see Eq. \ref{eq:critic}) 
along with the trained generator to generate a trajectory $z_0, z_1, . . . , z_T$ and %consider either $z_T$ or if %the $f(z_t)$ =\textit{'No'} as the final output. The iteration stops when 
%the critic generates "\textit{No}". 
stop when either $f(z_t)$ is generated by the generator or \textit{``No hint''} is generated by the critic. 
We also experimented with   \textit{chain of thought} prompting, where the generator generates a trajectory $z_0y_0, z_1y_1, . . . , z_Ty_T$ and stops when the critic generates \textit{``No hint''}. %\mete{Inference procedure is detailed in Algorithm \ref{alg:inference_refiner}.}
\if False
\begin{algorithm}\caption{Inference Iterative Critic-based Reasoning}
\begin{algorithmic}[1]
\State Initialize $answers  \gets$ empty list
\For {$i (batch) \gets 1$ to $N$}
    \State Initialize (reward) $r_i \gets 0$, $p_i \gets 1$
    \State Initialize (hint) $h_0, \hat{y}_{i,0} \gets No, []$
    \For {(turn) $t \gets 1$ to $T$}
        \State {$\hat{y}$ $\gets$  $\pi_{\theta}(y_i|c_i, h_{t-1}, \hat{y}_{i,t-1})$}
        \State {$h_{t} \gets \pi_{\beta}(c_i,  \hat{y}_{i})$}
            
        \If {$h_{t}$ == No}
                \State $answers$.append($\hat{y}$)
                \State break
        \EndIf
    \EndFor
    \State $answers$.append($\hat{y}$)
\EndFor
\State \Return $answers$
\end{algorithmic}
\end{algorithm}
\fi 

%In this section, we introduce our framework \ourmodel for generating intermediate hypothesis for natural language reasoning tasks. \ourmodel is inspired by the Actor-Critic Method \citep{pmlr-v48-mniha16}, where a policy function (or actor) returns a probability distribution over actions based on the given state and a critic (i.e., value function) evaluates the distribution. Instead of value function we propose to use textual feedback to update the generator model.  \ourmodel iteratively applies  two consecutive steps:  

\section{Experimental Setup} 
\begin{table}[t!]
\centering
\scalebox{0.7}{
\begin{tabular}{@{}l@{~~~~}c@{~~~}c@{~~~}}
\toprule
{\bf Generator Model} & {\bf Eq. ($z$)} & {\bf Ans. (${y}$)}\\\midrule
%\rowcolor{gray!25}
%\textbf{Full training data} &  &\\
UQA-base & {\cellcolor{NextBlue}34.1} & {\color{gray} --} \\
UQA-base + PPO & {\cellcolor{NextBlue}31.5} & {\color{gray} --} \\
\model$_{base}$ & \cellcolor{NextBlue}\textbf{47.2} & --\\
%\model$_{base}$ + {\color{orange} Oracle}& \cellcolor{NextBlue}{66.0} & {\color{gray} --} \\
%{\color{gray}} \\
\midrule
UQA-large & {\cellcolor{NextBlue}46.7} & {\color{gray} --} \\
UQA-large + PPO & \cellcolor{NextBlue}{48.2} & {\color{gray} --} \\
\model$_{large}$ & \cellcolor{NextBlue}\textbf{53.8} & {\color{gray} --} \\
%\model$_{large}$ + {\color{orange} Oracle} & \cellcolor{NextBlue}{68.1} & {\color{gray} --} \\
\midrule
%GPT$3.5$ & \cellcolor{NextBlue}{49.6} & {--} \\
%GPT3.5 & {--} & \cellcolor{NextBlue}{63} \\
%GPT$3.5$ + REFINER$_{critic}$ & \cellcolor{NextBlue}\textbf{59.6} & {--} \\
%GPT-$3.5$ + CoT & {59.3} & \cellcolor{Nextblush}{63.5} \\
%GPT3.5 + REFINER$_{critic}$ & \cellcolor{NextBlue}\textbf{59.6} & {--} \\
%GPT-$3.5$ + CoT + REFINER$_{critic}$ & {62.3} & \cellcolor{Nextblush}\textbf{66.4} \\
GPT-$3.5$ + CoT & {64.1} & \cellcolor{Nextblush}{67.1} \\
%GPT3.5 + REFINER$_{critic}$ & \cellcolor{NextBlue}\textbf{59.6} & {--} \\
GPT-$3.5$ + CoT + REFINER$_{critic}$ & \textbf{67.3} & \cellcolor{Nextblush}\textbf{70.6} \\

\bottomrule
\end{tabular}}
\caption{Results on MWP. 
% Eq.: Equation, Ans. Answer. 
Comparison of \ourmodel with baselines on the SVAMP dataset. The average score over three runs is reported (p<0.05). For models other than GPT-3.5, the answer can be obtained via symbolic execution of the equation and is thus a function of the validity of the equation.} %For GPT-3.5, the model is few-shot prompted to either generate the equation with variable names $z$, or generate the answer $y$.
%GPT-$3.5$: code-DaVinci-002,
\label{tab:results_mwp}
\end{table}

%In this section, we will describe .....
\textbf{Datasets.} We evaluate \ourmodel on three diverse tasks (examples in \Figref{fig:error_overview}). We briefly describe the datasets used for each task below.%\footnote{In Table \ref{tab:data_stat}, in the Appendix, we report the data statistics.}
\textit{Math Word Problem} (MWP): %is a mathematical reasoning task that evaluates the model's capability to reason over text to solve a math problem. 
We train our models on MAWPs \citep{koncel-kedziorski-etal-2016-mawps} dataset and evaluated our models on a challenging dataset SVAMP \citep{patel-etal-2021-nlp}. We evaluate our model on both the equation generation ($z$) and answer prediction ($y$) tasks. Similar to \citet{ling-etal-2017-program, amini-etal-2019-mathqa} for equation generation, we replace the numeric values with variable names, for example, $\texttt{number0}$, $\texttt{number1}$, \etc Further, we also evaluated on GSM8K \citep{cobbe2021gsm8k} dataset which consists of 8.5K high-quality linguistically diverse grade school math word problems.
For \textit{Synthetic Natural Language Reasoning} (sNLR), we use the dataset from \citet{liang2022holistic} with the difficulty level as hard. We evaluate our model on both inference rule generation ($z$) and consequent generation ($y$).
For \textit{Moral Story} (MS), we use a dataset from \citep{emelin-etal-2021-moral}, where we evaluate our model on moral norm  ${z}$ and the moral action ${y}$ generation.

\textbf{Training Details.} For each task, we train a UnifiedQa-\textsc{T5}-base model (UQA-base) \citep{khashabi-etal-2020-unifiedqa} as a critic (\S \ref{sec_3.1:critic_model}). 
For exploration (\S \ref{sec:3.2_generator}), we use nucleus sampling with $p = 0.5$. We select the hyper-parameters by the validation loss: for both the generator and critic model, we use the Adam optimizer with a learning rate of $1e^{-4}$. Each model is trained for $20$ epochs with early stopping based on validation loss. We trained all models on one A100 GPU. We run our models with $3$ random seeds and report the average results. For the human study, we selected outputs from the best models (baselines and our model) according to automatic metrics. We train models with $T=3$ iterations.

At inference time, we use greedy decoding for the generator and critic model with $T=1$ for the automatic critic and $T=3$ for the oracle critic. On the MWP and sNLR tasks, we use the exact match (EM) metric for intermediate steps (equation generation and inference rules) and accuracy (Acc) for the final answers. For MS, we conduct a manual evaluation study to assess the relevance of norms and moral actions\footnote{Since the automatic scores such as BLUE, ROUGE, etc. only account for word level similarity between gold norms or actions and generate norms or actions.}. 
Further evaluation details are provided in Appendix \ref{sec:Appendix_E}. 
To train the critic model, we used the feedback data generated using the rule-based perturbation strategy (see \S \ref{sec_3.1:critic_model}).

\textbf{Baselines.} We compare our method with three different LMs as generator models: UQA-base, UQA-large (supervised setting), %GPT-$3.5$-code-DaVinci-002  
GPT-$3.5$-\texttt{text-DaVinci-003} and ChatGPT (few-shot setting). We also compare \ourmodel to \textit{Proximal Policy Optimization} (PPO) RL-based method \citep{Schulman2017ProximalPO}. We use the implementation of PPO from \citep{Ramamurthy2022IsRL}.
%For GPT-3.5, the model is not finetuned but prompted with standard few-shot prompts \citep{NEURIPS2020_1457c0d6}, in which the LM is given in-context demonstrations of input-output pairs.
% before outputting a prediction for an inference-time instance. 
For GPT-$3.5$, we provide $2$ for demonstrations per class. We also experimented with \textit{chain of thought} (COT) prompting \citep{wei2022chain} where the model is prompted first to generate the intermediate steps ($z$) and then the final answer ($y$).  Note that the sNLR task is a synthetic task where the model needs to perform either one-hop or two-hop reasoning. \citet{10.5555/3491440.3491977} showed that fine-tuning large language models ($354$M parameter size) could achieve (99\% accuracy) high performance. Hence, we only compare our REFINER model with the UQA-base model ($220$M) (see Table \ref{tab:results_snlr}). %Note that sNLR is a synthetic task, hence for the supervised setting, we only report results with UQA-base. 
Since human annotation is expensive, we focus on comparing against the most meaningful baseline: UQA-large for MS task (see Table \ref{tab:results_mng}). It is important to highlight that our proposed framework is general, and one can use any other LMs as \textsc{generator} or \textsc{critic}.

\section{Results}\label{sec:result}

%\textbf{RQ1: Does \ourmodel improve the quality of the intermediate steps for diverse reasoning tasks? }
We evaluate our model on two aspects (i) performance on intermediate steps and (ii) performance on the final answer prediction.
Tables \ref{tab:results_mwp}, \ref{tab:results_snlr}, and \ref{tab:results_mng} show the performance comparisons. 
\begin{table}[t!]
\centering
\scalebox{0.7}{
\begin{tabular}{@{}l@{~~~}c@{~~~~}c@{~~~~}}
\toprule
{\bf Generator Model} & {\bf IR ($z$)} & {\bf Con (${y}$)}\\\midrule
%\rowcolor{gray!25}
%\textbf{Full training data} &  &\\
UQA-base & {90.6 $\pm$ 0.8} & 94.1 \\
\ourmodel$_{base}$ & \bf{93.5} $\pm$ 0.4 & \bf{97.3} \\
%\ourmodel$_{base}$ + \color{orange} Oracle & {96.2 $\pm$ 0.9} & {98.9}\\\hline
%\rowcolor{gray!25}
%\textbf{Few-Shot Setting} &  &\\
%GPT-$3.5$ + CoT & {17.4 $\pm$ 0.5} & {45.0} \\
%GPT-$3.5$ + CoT + \model  & \bf{26.8 $\pm$ 0.5} & \bf{46.6} \\
GPT-$3.5$ + CoT & {14.3 $\pm$ 0.9} & {40.6} \\
GPT-$3.5$ + CoT + \model  & \bf{21.1 $\pm$ 1.2} & \bf{42.1} \\
\bottomrule
\end{tabular}}
\caption{Results on sNLR task. The average score over three runs is reported (p<0.05). IR: Inference Rules (Exact Match), Con: Consequent (Accuracy)}
\label{tab:results_snlr}
\end{table}

\textbf{Performance on Intermediate Steps.}
Table \ref{tab:results_mwp} reports the performance of the MWP task. We explored two different scenarios: (i) where the model \colorbox{NextBlue}{only generates the equations} ($z$) with variable names replacing the numeric values, and (ii) where the model generates \colorbox{Nextblush}{both the equations and the final answers} together. We observe for both scenarios that \ourmodel significantly outperforms baseline models with comparable sizes. Notably, UQA-base benefits most ($+13.1$ EM) when adding a critic in the loop. We observe that GPT-$3.5$ significantly benefits from the \ourmodel trained critic. Since LLMs like GPT-$3.5$ ($175$B parameters) are expensive to finetune, the improvement in equation generation of $+3.2$ EM without any modification is important. Interestingly, we observe that GPT-$3.5$ + COT manages to have significantly higher accuracy in answer $y$ than in equation $z$ (see Table \ref{tab:results_mwp}). This result is similar to the observation made by \citet{ye2022the} and suggests that the intermediate equations can be unreliable. 
 %\mete{maybe here mention human-in-the-loop benefit?} 
Finally, \ourmodel could even outperform PPO, which uses BLEU-score as a reward function. This suggests that semi-structured fine-grained textual feedback is more beneficial than value-based (where values are from automatic metrics) reward feedback. Note that this result may vary when these models are optimized directly with complex human values, as shown in \citet{10.5555/3495724.3495977}. Qualitatively, \ourmodel can correct incorrect equations through structured feedback, fixing the operators within a multistep solution (see Fig. \ref{fig:example_multistep}). 

\begin{table}
  \centering
  \scalebox{0.8}{
  \begin{tabular}{p{1.5cm}cccc|cccc}
    \toprule
    \multirow{2}{1cm}{\textbf{}} & \multicolumn{4}{c}{\textbf{Norm ($z$)}} & \multicolumn{4}{c}{\textbf{Action ($y$)}}\\
    % \hline
    % \textbf{Inactive Modes} & \textbf{Description}\\
    \cline{2-9}
    \textbf{Model}& \textbf{I$\downarrow$} & \textbf{U$\downarrow$} & \textbf{R$\uparrow$} & $\alpha$ & \textbf{I$\downarrow$} & \textbf{U$\downarrow$} & \textbf{R$\uparrow$} & $\alpha$ \\
    %\hhline{~--}
    \hline
    B &  34 & 17 & 49 & 0.35 & 28 & 14 & 58 & 0.64 \\ 
    B+PPO & 38 & 10 & 52 & 0.38 & 31 & 17 & 52 & 0.38 \\ 
    \model & 19 & 12 & \bf{69} & 0.33 & 18 & 9 & \bf{73} & 0.55 \\ \bottomrule
  \end{tabular}}
  \caption{Results on Moral Norm and Moral Action. We report human evaluation. B: UQA-large; I: \textit{Irrelevant}, U: \textit{Unsure}; R: \textit{Relevant}; $\alpha$: Krippendorff's alpha}
  \label{tab:results_mng}
\end{table}
For sNLR, similar to \citet{liang2022holistic}, we observe that GPT-3.5 performs poorly (see Table \ref{tab:results_snlr}). %We also find that GPT-3.5 code-DaVinci is better than text-DaVinci on the sNLR task. 
\ourmodel improves $+2.9$, and $+6.8$ EM scores over UQA-base, and GPT-$3.5$, respectively. Contrary to the MWP, the final answer $y$ is not a symbolic execution away from the intermediate step $z$, but we still observe that \ourmodel focuses on improving the intermediate step $z$, resulting in significant improvements in the answer $y$ prediction. Again, we observe that \ourmodel with a UQA-base can outperform few-shot prompted GPT-$3.5$. Thus, our critic can identify the fine-grained reasoning errors and help improve the performance on inference rules generation.

For MS, we assess the generation quality with three human judges who indicate whether the generated norms and moral actions are relevant to the given moral story. Table \ref{tab:results_mng} summarises human evaluation results on $100$ moral story examples randomly sampled from the MS test dataset. More specifically, we report evaluation breakdown for both norm and moral action by the number of instances that are either \textit{Irrelevant}, \textit{Unsure} or \textit{Relevant} along with Krippendorf's $\alpha$ \cite{krippendorff} agreement scores. The results show an improvement of $20$ points, increasing the relevance over a strong UQA-large baseline. Hence, this suggests that a specialized critic model with $3$ times fewer parameters than the generator can improve the performance on generating reasoning steps.

\textbf{Performance on Final Answer Prediction.}
We observe that \model outperforms the strong LM baselines by $+3.5,+3.2,+15$ points for MWP, sNLR, and MS, respectively. These results support our hypothesis that generating better intermediate steps can result in better answer prediction. Notably, on the sNLR task, for GPT-$3.5$, we observe that by adding a critic, there is an improvement of +$6.8$ in inference step generation; however, only $+1.5$ in the consequent prediction. This result indicates that LLMs may either not use these intermediate steps to perform the deduction or fail to perform deduction.%, which leads to correct answers.  %these intermediate inferences can be inappropriate for the actual computation leading to the model's answer. Hence, training to improve the quality of the intermediate step can result in better performance on the final prediction. 

\begin{table}[t!]
\centering
\scalebox{0.7}
{
\begin{tabular}{@{}l@{~~}c@{~~}c@{~~}c@{~~}c@{~~}}
\toprule
\textbf{Generator Model} & \multicolumn{2}{c}{\textbf{SVAMP}} & \multicolumn{2}{c}{\textbf{GSM8K}}\\
& GPT-3.5 & ChatGPT & GPT-3.5 & ChatGPT \\
\midrule
CoT &  {67.1} & {68.2} & 63.5 & 74.1  \\
Self-reflection & {67.2} & 68.4 & 63.1 & 74.6 \\ 
Self-refine &  67.6 & 68.2 & 63.8 & 74.7\\
REFINER & \bf{70.6} & \bf{71.4} & \bf{66.2} & \bf{75.9} \\
\midrule
ReACT & 67.3 &  68.4 & 64.7 & 75.5\\ 
ReACT + REFINER  & \textbf{70.6} & \textbf{71.9} & \textbf{67.8} & \textbf{77.4} \\
\midrule
Self-consistency  & 69.5 & 70.4 & 65.5 &76.1 \\
Self-consistency + REFINER & \textbf{72.1} & \textbf{72.5} & \textbf{67.2} & \textbf{78.1} \\
\bottomrule
\end{tabular}}
\caption{\textbf{Comparison with different refinement methods} on SVAMP and GSM8K datasets. Averaged accuracy over three runs on the test sets is reported (p<0.05).}
\label{tab:sota_results}
\end{table}

\textbf{Comparing REFINER with other refinement methods.} 
In Table \ref{tab:sota_results}, we compare REFINER with two other recent refinement methods: Self-refine \citep{madaan2023selfrefine} and Self-reflection \citep{shinn2023reflexion} method on the SVAMP and GSM8K datasets. Both these baseline methods use LLMs to generate automatic feedback. Similar to \citet{madaan2023selfrefine}, we observe that self-refine has minor improvement for MWP tasks. On the contrary, we find that REFINER significantly improves the performance of GPT-3.5 and ChatGPT by +$3.3$ and +$2.2$ on SVAMP and GSM8K datasets, respectively. This highlights the benefit of training a \textit{specialised critic} that is grounded to the task. It can make LLMs more accurate than feedback from a general-purpose model (GPT-$3.5$ or ChatGPT).
In Appendix \S\ref{sec:compare_trained_training_free}, we have provided more details about the quality of feedback generated using our trained critic and GPT-$3.5$ (see Table \ref{tab:trained_critic}).
%\textbf{Can REFINER improve intermediate steps generated using Self-Consistency and ReACT methods?} 
Further, we assess the performance of REFINER in improving the CoT generated by two recent methods: Self-Consistency \citep{wang2023selfconsistency} and ReACT method \citep{yao2023react}.  We observe that REFINER can improve self-consistency and ReACT by +$2.02$ and +$2.9$. This demonstrates that a trained critic can be used as a \textit{tool} and can bring performance gains to different methods out-of-the-box (more details in Appendix \S \ref{sec:react}). 

\begin{table}[t!]
\centering
\scalebox{0.7}
{
\begin{tabular}{@{}l@{~~}c@{~~~~}}
\toprule
{\bf Model} & {\bf Eq. ($z$)}\\\midrule
%\rowcolor{gray!25}
%\textbf{SVAMP} - Equation Generation  \\
\model$_{base}$ + critic data$_{rule-based}$  & {47.2} \\
\model$_{base}$ - critic$_{inference}$ & {39.8} \\
\model$_{base}$ - critic$_{inference}$ - exp & {37.4}\\
\model$_{base}$ - critic$_{training}$ & {34.1}\\ 
\midrule
\model$_{base}$ + critic data$_{synthetic}$ & {44.1}\\ 
\model$_{base}$ + critic$_{Oracle}$ & 66.0 \\
\bottomrule
\end{tabular}}
\caption{\textbf{Ablation Result} on MWP task; Comparing model without critic during inference, and without the exploration (exp) phase during training. We report the exact match scores of the generated equation, comparable to \Tabref{tab:results_mwp}.}
\label{tab:ablation}
\end{table}
\textbf{Ablation.} To obtain better insight into the contributions of the individual components of our models, we perform an ablation study (Table \ref{tab:ablation}).
We observe that there is a considerable drop in performance from $47.2$ to $39.8$ when we do not use the critic model during inference. Hence, this result indicates that our generator model can leverage the feedback from the critic at inference time. Further, we find that the exploration step improves the performance $+3.3$ over the baseline model. This result supports our hypothesis that the exploration step increases the output variance and gives the generator model the opportunity to learn over a wide range of feedback. We compared the performance with the critic model trained on two different training data (see \S \ref{sec_3.1:critic_model}). We find that the critic trained on small automatically generated data using GPT-3.5 works better than without the critic in the loop. This result motivates researchers to use this method to generate negative samples to train their critic or preference learning model. Finally, we also observe that if the critic was perfect (Oracle), then \ourmodel can significantly improve the performance by fixing the mistakes generated by the generator model. This result indicates that \ourmodel can be seen as a framework that allows AI-AI and human-AI interaction.

\section{Analysis}\label{sec:analysis}
\begin{figure}[t]
  \centering
    \includegraphics[scale=1,height=4cm, width=0.35\paperwidth]{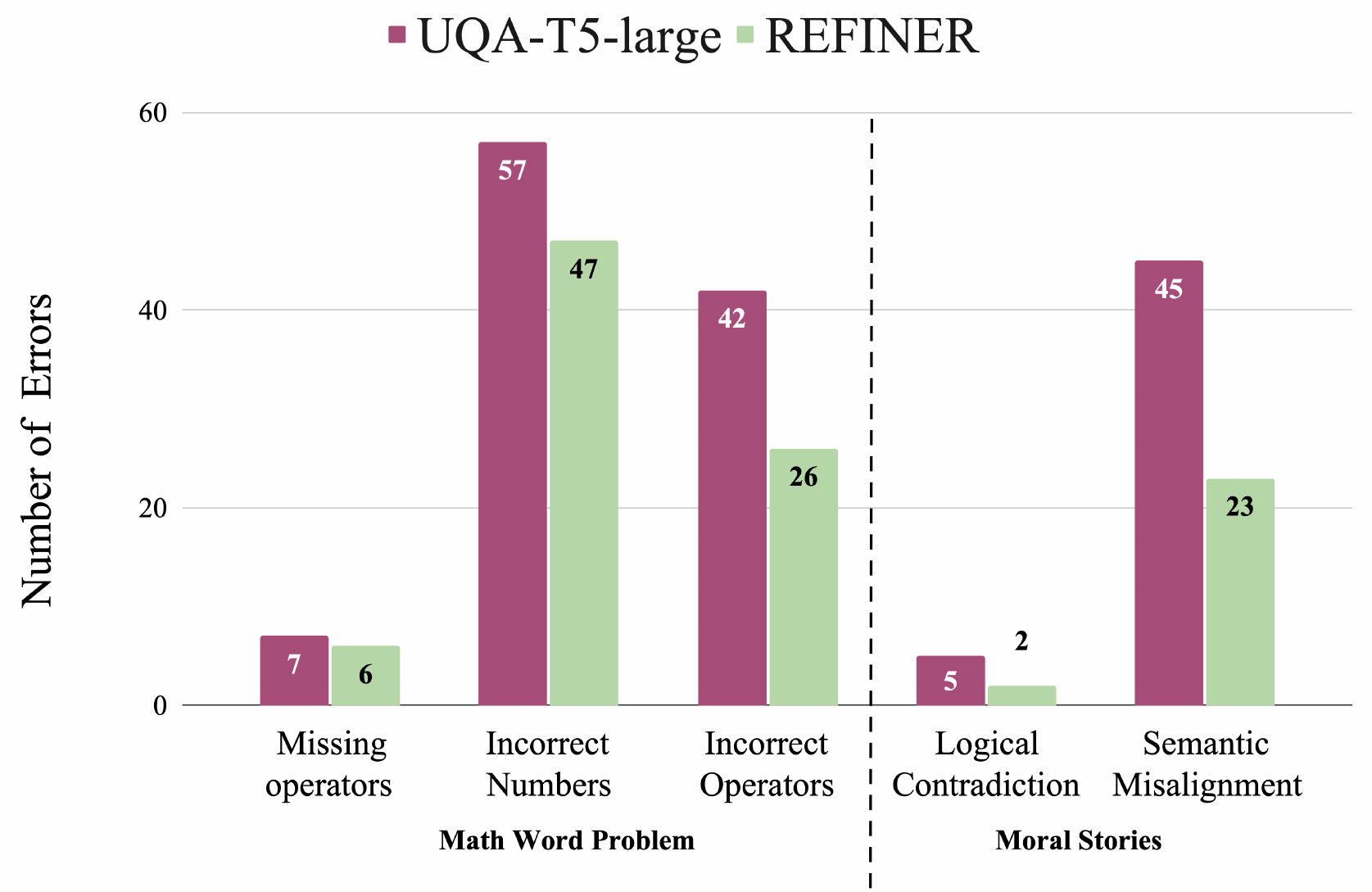}
    \caption{\textbf{Error analysis.} Number of errors made by baseline UQA-large and \ourmodel on 100 instances sampled randomly from test sets of both datasets. Errors are categorized according to Table \ref{tab:define_error_feedbacks}).}
    \label{fig:error_analysis}
\end{figure}

%\subsection{Quantitative Analysis}
\textbf{Error Analysis.}\label{sec:6.1_quantitative_analysis}
In order to get more insight into the performance of our method, we conduct a fine-grained error analysis on the MWP and MS datasets (Fig. \ref{fig:error_analysis}). We note that the most frequent errors are \textit{Incorrect Numbers} for MWP and \textit{Semantic Misalignment} for MS. An intuitive reason can be that for the MWP task, the models are sensitive to the numbers order as argued in \citep{patel-etal-2021-nlp}. For MS, generating norms grounded in the context is challenging. 
Our analyses show a clear trend that \ourmodel is able to considerably reduce the errors for both datasets. This indicates that our trained critic model could identify fine-grained reasoning errors during inference. 

\begin{figure}[t]
  \centering
    \includegraphics[scale=1,height=4cm, width=0.35\paperwidth]{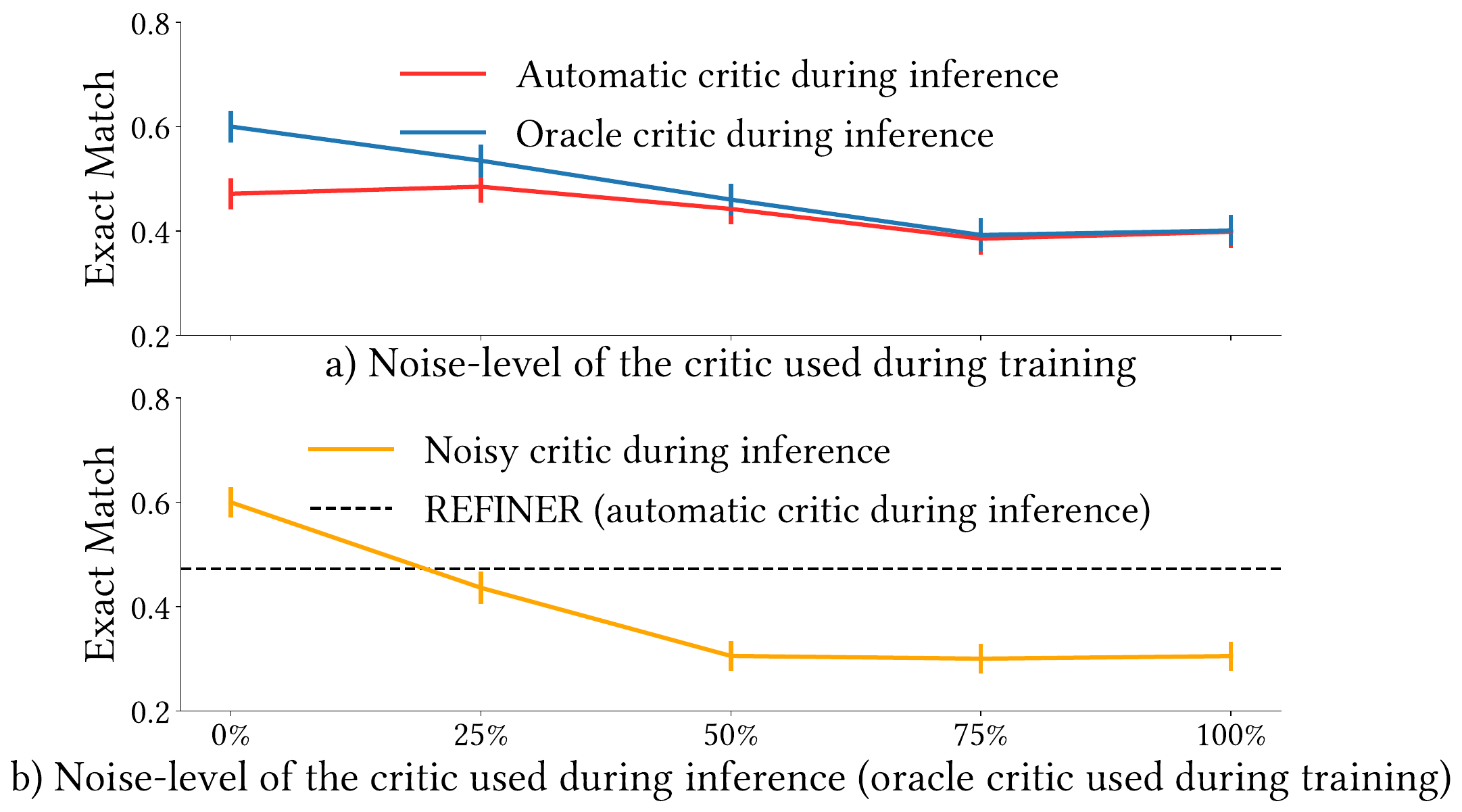}
    \caption{\textbf{Noisy-critics analysis}. In plot (a), we vary the noise level of the critic used during training ($0$ noise corresponds to oracle) and compare the resulting models when using the oracle and the training automatic critic during inference. In plot (b), we train with the oracle critic but vary the noise level of the critic used during inference.}
    \label{fig:noisy_analysis}
\end{figure}

\textbf{Noise Sensitivity.} 
To further understand the behaviour of the \ourmodel framework, we run variations with noisy critics for the MWP task. We replace the oracle critic used during training with a noisy critic in (\Figref{fig:noisy_analysis}~(a)) to inspect how training with an imperfect critic impacts the generator. We also use a noisy critic at inference while keep the oracle critic during training (in \Figref{fig:noisy_analysis}~(b)). %This analysis is performed on the SVAMP dataset for MWP. 
The noisy critics are generated by random perturbations of the oracle critic; for a noise-level $\epsilon$, the oracle feedback is replaced by random feedback with probability $\epsilon$.

Fig. \ref{fig:noisy_analysis}~(a) shows that when training with a very noisy critic ($>75\%$ noise), the generator LM learns to ignore the critic, as there is no difference between using the trained critic or the oracle during inference. Interestingly, training with a bit of noise ($<50\%$) does not seem to harm the model, as performances are not statistically different than training with the oracle (noise of $0\%$).
Fig. \ref{fig:noisy_analysis}~(b) depicts the quality of the critic used at inference time has a huge impact. Having oracle provide feedback is by far the best scenario. Already with $25\%$ noise, the critic makes the generator perform worse than using our trained critic (\ourmodel). With more than $50\%$ noise, the critic significantly harms the generator. The generator, trained with an oracle critic, has learned to trust the critic and expects useful feedback.

\textbf{Qualitative Analysis.} To explain the findings in \S\ref{sec:6.1_quantitative_analysis}, we further manually analyze 100 instances for the MWP task. We observe two different scenarios when REFINER failed to fix the outputs generated by \textsc{generator} model: (a) when the \textsc{critic} model provides a \textit{correct} feedback; however, the \textsc{generator} model still generates \textit{incorrect} equation, and (b) the \textsc{critic} model provides an \textit{incomplete} or \textit{partially correct} feedback. The former case indicates that either the \textsc{generator} model makes mistakes in following the instruction from the \textsc{critic} or the feedback from the critic can be ambiguous. For example, in Appx \Figref{fig:qualitative_analysis}, (b) we observe the case when the critic is correct, but the feedback could result in an incorrect equation. The latter case indicates that our trained critic model generates incorrect feedback, which can result in incorrect or partially correct equations. We also observe that our \textsc{critic} model failed to generate correct feedback when the \textsc{generator} model generates incorrect equations with multiple mistakes.

\begin{table}[t!]
\centering
\scalebox{0.7}
{
\begin{tabular}{@{}l@{~~}c@{~~~~}c@{~~~~}c@{~~~~}c@{~~~~}}
\toprule
{\bf Task} & {\bf UQA (220M)} & {\bf UQA (770M)} & {\bf GPT-3 (175B)}  \\\midrule
%\rowcolor{gray!25}
%\textbf{SVAMP} - Equation Generation  \\
MWP  & 69.5 +/- 2.6 & 73.4 +/- 3.7 & 63.5 +/- 5.6\\
sNLR & 95.5 +/- 1.4 & 98 +/- 2.2 & 34.5 +/- 2.4\\ 
MN & 77.4 +/-2.5 & 80 +/- 4.5 & 76.4 +/-3.5\\
\bottomrule
\end{tabular}}
\caption{\textbf{Comparing the performance of different critic models}. Exact-match score is reported.}
%\vspace{-1.5em}
\label{tab:trained_critic}
\end{table}

\textbf{Quality of the feedback.}\label{sec:compare_trained_training_free} To better understand the difference in the quality of the feedback, we compare our trained critic model with GPT-3.5.  We assess the quality of the feedback on 500 instances per task and report the exact match scores in Table \ref{tab:trained_critic}. Please note that we include instances where the critic feedback should say the solution is correct and hence generate 'No'. For GPT-3.5, we have provided (two) few-shot examples per type of error and two examples with 'No' as feedback. Our results show that trained critic (UQA) can comprehensively outperform GPT-3.5. We observe that GPT-3.5 performs well in identifying when the answer is correct. However, it makes errors when asked to generate meaningful semi-structured feedback for incorrect reasoning steps.

\section{Conclusion} 

In this paper, we propose \ourmodel, a framework to improve the reasoning abilities of LMs through an iterative feedback loop between two models, a \textit{generator} and a \textit{critic}. Our evaluation of this framework on three reasoning tasks showed structured and fine-grained feedback on intermediate reasoning errors results in significant performance gains, surpassing scalar value feedback. Our trained critic model alone, even when noisy, can improve intermediate representations of LMs, showing that \ourmodel can significantly boost LMs' performance on reasoning tasks. Our REFINER framework is very general and, in principle, might be applied to steer language models in performing different reasoning tasks. More specifically, the \textit{critic} model can be seen as a tool for LLMs to refine their generation quality.

\section*{Acknowledgment}

We would like to thank Martin Josifoski, Syrielle Montariol, and Zeming Chen for their helpful feedback on a draft version of the paper. We acknowledge the support of the ICT-48 Network of AI Research Excellence Center “TAILOR” (EU Horizon 2020, GA No 952215). West's lab is partly supported by grants from the Swiss National Science Foundation (200021\_185043), Swiss Data Science Center (P22\_08), H2020 (952215), Microsoft Swiss Joint Research Center, and Google, and by generous gifts from Facebook, Google, and Microsoft. Antoine Bosselut gratefully acknowledges the support of Innosuisse under PFFS-21-29, the EPFL Science Seed Fund, the EPFL Center for Imaging, Sony Group Corporation, and the Allen Institute for AI.

\newpage

\section*{Limitations}
Our \ourmodel framework could not be comprehensively evaluated on all applicable downstream reasoning tasks due to their sheer number. While deliberately distinct, we focused on only three different reasoning tasks in order to study how natural language reasoning feedback can impact downstream tasks. We believe this represents an initial but important step towards exploring automated natural language feedback on intermediate representations.
In addition, the critic we presented here is specific for each task, while the ideal critic would be a general one, capable of providing feedback on a wide range of reasoning tasks. Similarly, we considered fine-grained reasoning errors specific to each reasoning task. Recent work has mentioned several other fine-grained reasoning errors \citep{anonymous2023roscoe}, which can't be fully covered by the reasoning tasks we considered. Generalizing both the critic and fine-grained error types emerges as both the main limitations of this paper and the directions of future work. Finally, with LLMs being deployed more and more for real-life applications (medical domain, making important decisions), we believe it is crucial to develop expert models and automatic feedback mechanisms to inspect model generations and improve them. LLMs are impressive and work well on several NLP tasks, but they are not expert systems. Our work aims to address this gap by showing that adding interventions/feedback from critics (specialised finetuned critics) can help the LLM model to be more accurate—additionally, making the whole process more transparent.

%Finally, this study shows how improving the intermediate representation can impact the final reasoning answers. However, this work doesn't answer how language models utilise these intermediate representations to improve their performance or not.}

%\mete{Firstly, although diverse in their nature, we have considered only three reasoning tasks to evaluate \ourmodel on among which only one task (MS) requires general commonsense reasoning. The open-ended nature of this type of reasoning makes it harder to provide structured feedback which \ourmodel framework relies on. Although, with MS task, we have shown one successful application of our method in this form of unconstrained setting, effectiveness of \ourmodel on other types of reasoning scenarios will have to be shown. Lastly, in this work, we only explore a few fine-grained error types per task. As described in \cite{anonymous2023roscoe}, there are a wide variety of potential reasoning errors that could be considered for the evaluation of our method, however, we also believe, our framework is general enough to be adapted easily for other types of reasoning errors in the future.}

\section*{Ethical Considerations}
In this paper, we experiment with existing datasets which are, to the best of our knowledge, adequately cited. Our proposed framework \ourmodel is designed to improve the reasoning abilities of LMs. These LMs have been shown to encode biases about race, gender, and many other demographic attributes \cite{ethical-social-risks}, \cite{sheng-etal-2020-towards}. Since our framework does not offer a way to mitigate these biases, models improved using this framework could still reflect the same harmful behaviours normally exhibited by these models. We recommend anyone deploying our model \textit{off-the-shelf} should first check whether the model is harmful towards any protected group, and appropriate mitigation should be taken. In addition, our MS task is based on a dataset of situations, intentions, and actions that heavily skew towards Western culture and social norms \cite{emelin-etal-2021-moral}. Consequently, our human evaluation on the MS task was done with AMT workers based in the US who were paid adequately for the average time it took to solve the task.

\bibliography{anthology,custom}

\begin{thebibliography}{63}
\expandafter\ifx\csname natexlab\endcsname\relax\def\natexlab#1{#1}\fi

\bibitem[{Amini et~al.(2019)Amini, Gabriel, Lin, Koncel-Kedziorski, Choi, and Hajishirzi}]{amini-etal-2019-mathqa}
Aida Amini, Saadia Gabriel, Shanchuan Lin, Rik Koncel-Kedziorski, Yejin Choi, and Hannaneh Hajishirzi. 2019.
\newblock \href {https://doi.org/10.18653/v1/N19-1245} {{M}ath{QA}: Towards interpretable math word problem solving with operation-based formalisms}.
\newblock In \emph{Proceedings of the 2019 Conference of the North {A}merican Chapter of the Association for Computational Linguistics: Human Language Technologies, Volume 1 (Long and Short Papers)}, pages 2357--2367, Minneapolis, Minnesota. Association for Computational Linguistics.

\bibitem[{Andor et~al.(2019)Andor, He, Lee, and Pitler}]{andor-etal-2019-giving}
Daniel Andor, Luheng He, Kenton Lee, and Emily Pitler. 2019.
\newblock \href {https://doi.org/10.18653/v1/D19-1609} {Giving {BERT} a calculator: Finding operations and arguments with reading comprehension}.
\newblock In \emph{Proceedings of the 2019 Conference on Empirical Methods in Natural Language Processing and the 9th International Joint Conference on Natural Language Processing (EMNLP-IJCNLP)}, pages 5947--5952, Hong Kong, China. Association for Computational Linguistics.

\bibitem[{Austin et~al.(2021)Austin, Odena, Nye, Bosma, Michalewski, Dohan, Jiang, Cai, Terry, Le, and Sutton}]{austin2021programsynth}
Jacob Austin, Augustus Odena, Maxwell Nye, Maarten Bosma, Henryk Michalewski, David Dohan, Ellen Jiang, Carrie Cai, Michael Terry, Quoc Le, and Charles Sutton. 2021.
\newblock \href {https://doi.org/10.48550/ARXIV.2108.07732} {Program synthesis with large language models}.

\bibitem[{Bai et~al.(2022)Bai, Kadavath, Kundu, Askell, Kernion, Jones, Chen, Goldie, Mirhoseini, McKinnon, Chen, Olsson, Olah, Hernandez, Drain, Ganguli, Li, Tran-Johnson, Perez, Kerr, Mueller, Ladish, Landau, Ndousse, Lukosuite, Lovitt, Sellitto, Elhage, Schiefer, Mercado, DasSarma, Lasenby, Larson, Ringer, Johnston, Kravec, Showk, Fort, Lanham, Telleen-Lawton, Conerly, Henighan, Hume, Bowman, Hatfield-Dodds, Mann, Amodei, Joseph, McCandlish, Brown, and Kaplan}]{bai2022constitutional}
Yuntao Bai, Saurav Kadavath, Sandipan Kundu, Amanda Askell, Jackson Kernion, Andy Jones, Anna Chen, Anna Goldie, Azalia Mirhoseini, Cameron McKinnon, Carol Chen, Catherine Olsson, Christopher Olah, Danny Hernandez, Dawn Drain, Deep Ganguli, Dustin Li, Eli Tran-Johnson, Ethan Perez, Jamie Kerr, Jared Mueller, Jeffrey Ladish, Joshua Landau, Kamal Ndousse, Kamile Lukosuite, Liane Lovitt, Michael Sellitto, Nelson Elhage, Nicholas Schiefer, Noemi Mercado, Nova DasSarma, Robert Lasenby, Robin Larson, Sam Ringer, Scott Johnston, Shauna Kravec, Sheer~El Showk, Stanislav Fort, Tamera Lanham, Timothy Telleen-Lawton, Tom Conerly, Tom Henighan, Tristan Hume, Samuel~R. Bowman, Zac Hatfield-Dodds, Ben Mann, Dario Amodei, Nicholas Joseph, Sam McCandlish, Tom Brown, and Jared Kaplan. 2022.
\newblock \href {https://doi.org/10.48550/ARXIV.2212.08073} {Constitutional ai: Harmlessness from ai feedback}.

\bibitem[{Brown et~al.(2020)Brown, Mann, Ryder, Subbiah, Kaplan, Dhariwal, Neelakantan, Shyam, Sastry, Askell, Agarwal, Herbert-Voss, Krueger, Henighan, Child, Ramesh, Ziegler, Wu, Winter, Hesse, Chen, Sigler, Litwin, Gray, Chess, Clark, Berner, McCandlish, Radford, Sutskever, and Amodei}]{NEURIPS2020_1457c0d6}
Tom Brown, Benjamin Mann, Nick Ryder, Melanie Subbiah, Jared~D Kaplan, Prafulla Dhariwal, Arvind Neelakantan, Pranav Shyam, Girish Sastry, Amanda Askell, Sandhini Agarwal, Ariel Herbert-Voss, Gretchen Krueger, Tom Henighan, Rewon Child, Aditya Ramesh, Daniel Ziegler, Jeffrey Wu, Clemens Winter, Chris Hesse, Mark Chen, Eric Sigler, Mateusz Litwin, Scott Gray, Benjamin Chess, Jack Clark, Christopher Berner, Sam McCandlish, Alec Radford, Ilya Sutskever, and Dario Amodei. 2020.
\newblock \href {https://proceedings.neurips.cc/paper/2020/file/1457c0d6bfcb4967418bfb8ac142f64a-Paper.pdf} {Language models are few-shot learners}.
\newblock In \emph{Advances in Neural Information Processing Systems}, volume~33, pages 1877--1901. Curran Associates, Inc.

\bibitem[{Christiano et~al.(2017)Christiano, Leike, Brown, Martic, Legg, and Amodei}]{christiano2017deepRL}
Paul Christiano, Jan Leike, Tom~B. Brown, Miljan Martic, Shane Legg, and Dario Amodei. 2017.
\newblock \href {https://doi.org/10.48550/ARXIV.1706.03741} {Deep reinforcement learning from human preferences}.

\bibitem[{Clark et~al.(2021)Clark, Tafjord, and Richardson}]{10.5555/3491440.3491977}
Peter Clark, Oyvind Tafjord, and Kyle Richardson. 2021.
\newblock Transformers as soft reasoners over language.
\newblock In \emph{Proceedings of the Twenty-Ninth International Joint Conference on Artificial Intelligence}, IJCAI'20.

\bibitem[{Cobbe et~al.(2021{\natexlab{a}})Cobbe, Kosaraju, Bavarian, Chen, Jun, Kaiser, Plappert, Tworek, Hilton, Nakano, Hesse, and Schulman}]{cobbe2021verifiers}
Karl Cobbe, Vineet Kosaraju, Mohammad Bavarian, Mark Chen, Heewoo Jun, Lukasz Kaiser, Matthias Plappert, Jerry Tworek, Jacob Hilton, Reiichiro Nakano, Christopher Hesse, and John Schulman. 2021{\natexlab{a}}.
\newblock \href {https://doi.org/10.48550/ARXIV.2110.14168} {Training verifiers to solve math word problems}.

\bibitem[{Cobbe et~al.(2021{\natexlab{b}})Cobbe, Kosaraju, Bavarian, Chen, Jun, Kaiser, Plappert, Tworek, Hilton, Nakano, Hesse, and Schulman}]{cobbe2021gsm8k}
Karl Cobbe, Vineet Kosaraju, Mohammad Bavarian, Mark Chen, Heewoo Jun, Lukasz Kaiser, Matthias Plappert, Jerry Tworek, Jacob Hilton, Reiichiro Nakano, Christopher Hesse, and John Schulman. 2021{\natexlab{b}}.
\newblock Training verifiers to solve math word problems.
\newblock \emph{arXiv preprint arXiv:2110.14168}.

\bibitem[{Elgohary et~al.(2020)Elgohary, Hosseini, and Hassan~Awadallah}]{elgohary-etal-2020-speak}
Ahmed Elgohary, Saghar Hosseini, and Ahmed Hassan~Awadallah. 2020.
\newblock \href {https://doi.org/10.18653/v1/2020.acl-main.187} {Speak to your parser: Interactive text-to-{SQL} with natural language feedback}.
\newblock In \emph{Proceedings of the 58th Annual Meeting of the Association for Computational Linguistics}, pages 2065--2077, Online. Association for Computational Linguistics.

\bibitem[{Elgohary et~al.(2021)Elgohary, Meek, Richardson, Fourney, Ramos, and Awadallah}]{elgohary-etal-2021-nl}
Ahmed Elgohary, Christopher Meek, Matthew Richardson, Adam Fourney, Gonzalo Ramos, and Ahmed~Hassan Awadallah. 2021.
\newblock \href {https://doi.org/10.18653/v1/2021.naacl-main.444} {{NL}-{EDIT}: Correcting semantic parse errors through natural language interaction}.
\newblock In \emph{Proceedings of the 2021 Conference of the North American Chapter of the Association for Computational Linguistics: Human Language Technologies}, pages 5599--5610, Online. Association for Computational Linguistics.

\bibitem[{Emelin et~al.(2021)Emelin, Le~Bras, Hwang, Forbes, and Choi}]{emelin-etal-2021-moral}
Denis Emelin, Ronan Le~Bras, Jena~D. Hwang, Maxwell Forbes, and Yejin Choi. 2021.
\newblock \href {https://doi.org/10.18653/v1/2021.emnlp-main.54} {Moral stories: Situated reasoning about norms, intents, actions, and their consequences}.
\newblock In \emph{Proceedings of the 2021 Conference on Empirical Methods in Natural Language Processing}, pages 698--718, Online and Punta Cana, Dominican Republic. Association for Computational Linguistics.

\bibitem[{Feng et~al.(2021)Feng, Gangal, Wei, Chandar, Vosoughi, Mitamura, and Hovy}]{feng-etal-2021-survey}
Steven~Y. Feng, Varun Gangal, Jason Wei, Sarath Chandar, Soroush Vosoughi, Teruko Mitamura, and Eduard Hovy. 2021.
\newblock \href {https://doi.org/10.18653/v1/2021.findings-acl.84} {A survey of data augmentation approaches for {NLP}}.
\newblock In \emph{Findings of the Association for Computational Linguistics: ACL-IJCNLP 2021}, pages 968--988, Online. Association for Computational Linguistics.

\bibitem[{Geva et~al.(2020)Geva, Gupta, and Berant}]{geva-etal-2020-injecting}
Mor Geva, Ankit Gupta, and Jonathan Berant. 2020.
\newblock \href {https://doi.org/10.18653/v1/2020.acl-main.89} {Injecting numerical reasoning skills into language models}.
\newblock In \emph{Proceedings of the 58th Annual Meeting of the Association for Computational Linguistics}, pages 946--958, Online. Association for Computational Linguistics.

\bibitem[{Golovneva et~al.(2022)Golovneva, Chen, Poff, Corredor, Zettlemoyer, Fazel-Zarandi, and Celikyilmaz}]{golovneva2022roscoe}
Olga Golovneva, Moya Chen, Spencer Poff, Martin Corredor, Luke Zettlemoyer, Maryam Fazel-Zarandi, and Asli Celikyilmaz. 2022.
\newblock \href {https://doi.org/10.48550/ARXIV.2212.07919} {Roscoe: A suite of metrics for scoring step-by-step reasoning}.

\bibitem[{Golovneva et~al.(2023)Golovneva, Chen, Poff, Corredor, Zettlemoyer, Fazel-Zarandi, and Celikyilmaz}]{anonymous2023roscoe}
Olga Golovneva, Moya Chen, Spencer Poff, Martin Corredor, Luke Zettlemoyer, Maryam Fazel-Zarandi, and Asli Celikyilmaz. 2023.
\newblock \href {https://openreview.net/forum?id=xYlJRpzZtsY} {{ROSCOE}: A suite of metrics for scoring step-by-step reasoning}.
\newblock In \emph{The Eleventh International Conference on Learning Representations}.

\bibitem[{Hedderich et~al.(2021)Hedderich, Lange, Adel, Str{\"o}tgen, and Klakow}]{hedderich-etal-2021-survey}
Michael~A. Hedderich, Lukas Lange, Heike Adel, Jannik Str{\"o}tgen, and Dietrich Klakow. 2021.
\newblock \href {https://doi.org/10.18653/v1/2021.naacl-main.201} {A survey on recent approaches for natural language processing in low-resource scenarios}.
\newblock In \emph{Proceedings of the 2021 Conference of the North American Chapter of the Association for Computational Linguistics: Human Language Technologies}, pages 2545--2568, Online. Association for Computational Linguistics.

\bibitem[{Huang et~al.(2022)Huang, Gu, Hou, Wu, Wang, Yu, and Han}]{huang2022selfimprove}
Jiaxin Huang, Shixiang~Shane Gu, Le~Hou, Yuexin Wu, Xuezhi Wang, Hongkun Yu, and Jiawei Han. 2022.
\newblock \href {https://doi.org/10.48550/ARXIV.2210.11610} {Large language models can self-improve}.

\bibitem[{Jie et~al.(2022)Jie, Li, and Lu}]{jie-etal-2022-learning}
Zhanming Jie, Jierui Li, and Wei Lu. 2022.
\newblock \href {https://doi.org/10.18653/v1/2022.acl-long.410} {Learning to reason deductively: Math word problem solving as complex relation extraction}.
\newblock In \emph{Proceedings of the 60th Annual Meeting of the Association for Computational Linguistics (Volume 1: Long Papers)}, pages 5944--5955, Dublin, Ireland. Association for Computational Linguistics.

\bibitem[{Khashabi et~al.(2020)Khashabi, Min, Khot, Sabharwal, Tafjord, Clark, and Hajishirzi}]{khashabi-etal-2020-unifiedqa}
Daniel Khashabi, Sewon Min, Tushar Khot, Ashish Sabharwal, Oyvind Tafjord, Peter Clark, and Hannaneh Hajishirzi. 2020.
\newblock \href {https://doi.org/10.18653/v1/2020.findings-emnlp.171} {{UNIFIEDQA}: Crossing format boundaries with a single {QA} system}.
\newblock In \emph{Findings of the Association for Computational Linguistics: EMNLP 2020}, pages 1896--1907, Online. Association for Computational Linguistics.

\bibitem[{Kiehne et~al.(2022)Kiehne, Kroll, and Balke}]{kiehne-emnlp-2022}
Niklas Kiehne, Hermann Kroll, and Wolf-Tilo Balke. 2022.
\newblock Contextualizing language models for norms diverging from social majority.
\newblock In \emph{Findings of the EMNLP 2022}, Abu Dhabi, United Arab Emirates. Association for Computational Linguistics, Association for Computational Linguistics.

\bibitem[{Kim et~al.(2022)Kim, Ki, Rhim, and Gweon}]{kim-etal-2022-ept}
Bugeun Kim, Kyung~Seo Ki, Sangkyu Rhim, and Gahgene Gweon. 2022.
\newblock \href {https://doi.org/10.18653/v1/2022.acl-long.305} {{EPT}-{X}: An expression-pointer transformer model that generates e{X}planations for numbers}.
\newblock In \emph{Proceedings of the 60th Annual Meeting of the Association for Computational Linguistics (Volume 1: Long Papers)}, pages 4442--4458, Dublin, Ireland. Association for Computational Linguistics.

\bibitem[{Koncel-Kedziorski et~al.(2016)Koncel-Kedziorski, Roy, Amini, Kushman, and Hajishirzi}]{koncel-kedziorski-etal-2016-mawps}
Rik Koncel-Kedziorski, Subhro Roy, Aida Amini, Nate Kushman, and Hannaneh Hajishirzi. 2016.
\newblock \href {https://doi.org/10.18653/v1/N16-1136} {{MAWPS}: A math word problem repository}.
\newblock In \emph{Proceedings of the 2016 Conference of the North {A}merican Chapter of the Association for Computational Linguistics: Human Language Technologies}, pages 1152--1157, San Diego, California. Association for Computational Linguistics.

\bibitem[{Krippendorff(2018)}]{krippendorff}
Klaus Krippendorff. 2018.
\newblock \emph{Content analysis: An introduction to its methodology.}
\newblock Sage Publications.

\bibitem[{Lampinen et~al.(2022)Lampinen, Roy, Dasgupta, Chan, Tam, McClelland, Yan, Santoro, Rabinowitz, Wang et~al.}]{lampinen2022tell}
Andrew~K Lampinen, Nicholas~A Roy, Ishita Dasgupta, Stephanie~CY Chan, Allison~C Tam, James~L McClelland, Chen Yan, Adam Santoro, Neil~C Rabinowitz, Jane~X Wang, et~al. 2022.
\newblock Tell me why! explanations support learning relational and causal structure.
\newblock In \emph{International Conference on Machine Learning}.

\bibitem[{Liang et~al.(2022)Liang, Bommasani, Lee, Tsipras, Soylu, Yasunaga, Zhang, Narayanan, Wu, Kumar et~al.}]{liang2022holistic}
Percy Liang, Rishi Bommasani, Tony Lee, Dimitris Tsipras, Dilara Soylu, Michihiro Yasunaga, Yian Zhang, Deepak Narayanan, Yuhuai Wu, Ananya Kumar, et~al. 2022.
\newblock Holistic evaluation of language models.
\newblock \emph{arXiv preprint arXiv:2211.09110}.

\bibitem[{Ling et~al.(2017)Ling, Yogatama, Dyer, and Blunsom}]{ling-etal-2017-program}
Wang Ling, Dani Yogatama, Chris Dyer, and Phil Blunsom. 2017.
\newblock \href {https://doi.org/10.18653/v1/P17-1015} {Program induction by rationale generation: Learning to solve and explain algebraic word problems}.
\newblock In \emph{Proceedings of the 55th Annual Meeting of the Association for Computational Linguistics (Volume 1: Long Papers)}, pages 158--167, Vancouver, Canada. Association for Computational Linguistics.

\bibitem[{Madaan et~al.(2023)Madaan, Tandon, Gupta, Hallinan, Gao, Wiegreffe, Alon, Dziri, Prabhumoye, Yang, Welleck, Majumder, Gupta, Yazdanbakhsh, and Clark}]{madaan2023selfrefine}
Aman Madaan, Niket Tandon, Prakhar Gupta, Skyler Hallinan, Luyu Gao, Sarah Wiegreffe, Uri Alon, Nouha Dziri, Shrimai Prabhumoye, Yiming Yang, Sean Welleck, Bodhisattwa~Prasad Majumder, Shashank Gupta, Amir Yazdanbakhsh, and Peter Clark. 2023.
\newblock \href {http://arxiv.org/abs/2303.17651} {Self-refine: Iterative refinement with self-feedback}.

\bibitem[{Marasovic et~al.(2022)Marasovic, Beltagy, Downey, and Peters}]{marasovic-etal-2022-shot}
Ana Marasovic, Iz~Beltagy, Doug Downey, and Matthew Peters. 2022.
\newblock \href {https://doi.org/10.18653/v1/2022.findings-naacl.31} {Few-shot self-rationalization with natural language prompts}.
\newblock In \emph{Findings of the Association for Computational Linguistics: NAACL 2022}, pages 410--424, Seattle, United States. Association for Computational Linguistics.

\bibitem[{Martin et~al.(2022)Martin, Quispe, Ollion, Le~Corff, Strub, and Pietquin}]{martin-etal-2022-learning}
Alice Martin, Guillaume Quispe, Charles Ollion, Sylvain Le~Corff, Florian Strub, and Olivier Pietquin. 2022.
\newblock \href {https://doi.org/10.18653/v1/2022.naacl-main.2} {Learning natural language generation with truncated reinforcement learning}.
\newblock In \emph{Proceedings of the 2022 Conference of the North American Chapter of the Association for Computational Linguistics: Human Language Technologies}, pages 12--37, Seattle, United States. Association for Computational Linguistics.

\bibitem[{Mehta and Goldwasser(2019)}]{mehta-goldwasser-2019-improving}
Nikhil Mehta and Dan Goldwasser. 2019.
\newblock \href {https://doi.org/10.18653/v1/N19-1195} {Improving natural language interaction with robots using advice}.
\newblock In \emph{Proceedings of the 2019 Conference of the North {A}merican Chapter of the Association for Computational Linguistics: Human Language Technologies, Volume 1 (Long and Short Papers)}, pages 1962--1967, Minneapolis, Minnesota. Association for Computational Linguistics.

\bibitem[{Nguyen et~al.(2021)Nguyen, Misra, Schapire, Dudík, and Shafto}]{nguyen2021interactive}
Khanh Nguyen, Dipendra Misra, Robert Schapire, Miro Dudík, and Patrick Shafto. 2021.
\newblock \href {https://doi.org/10.48550/ARXIV.2102.07024} {Interactive learning from activity description}.

\bibitem[{Nye et~al.(2021)Nye, Andreassen, Gur-Ari, Michalewski, Austin, Bieber, Dohan, Lewkowycz, Bosma, Luan, Sutton, and Odena}]{nye2021scratchpads}
Maxwell Nye, Anders~Johan Andreassen, Guy Gur-Ari, Henryk Michalewski, Jacob Austin, David Bieber, David Dohan, Aitor Lewkowycz, Maarten Bosma, David Luan, Charles Sutton, and Augustus Odena. 2021.
\newblock \href {https://doi.org/10.48550/ARXIV.2112.00114} {Show your work: Scratchpads for intermediate computation with language models}.

\bibitem[{Patel et~al.(2021)Patel, Bhattamishra, and Goyal}]{patel-etal-2021-nlp}
Arkil Patel, Satwik Bhattamishra, and Navin Goyal. 2021.
\newblock \href {https://doi.org/10.18653/v1/2021.naacl-main.168} {Are {NLP} models really able to solve simple math word problems?}
\newblock In \emph{Proceedings of the 2021 Conference of the North American Chapter of the Association for Computational Linguistics: Human Language Technologies}, pages 2080--2094, Online. Association for Computational Linguistics.

\bibitem[{Paul and Frank(2021)}]{paul-frank-2021-coins}
Debjit Paul and Anette Frank. 2021.
\newblock \href {https://doi.org/10.18653/v1/2021.acl-long.395} {{COINS}: Dynamically generating {CO}ntextualized inference rules for narrative story completion}.
\newblock In \emph{Proceedings of the 59th Annual Meeting of the Association for Computational Linguistics and the 11th International Joint Conference on Natural Language Processing (Volume 1: Long Papers)}, pages 5086--5099, Online. Association for Computational Linguistics.

\bibitem[{Pi{\k{e}}kos et~al.(2021)Pi{\k{e}}kos, Malinowski, and Michalewski}]{piekos-etal-2021-measuring}
Piotr Pi{\k{e}}kos, Mateusz Malinowski, and Henryk Michalewski. 2021.
\newblock \href {https://doi.org/10.18653/v1/2021.acl-short.49} {Measuring and improving {BERT}{'}s mathematical abilities by predicting the order of reasoning.}
\newblock In \emph{Proceedings of the 59th Annual Meeting of the Association for Computational Linguistics and the 11th International Joint Conference on Natural Language Processing (Volume 2: Short Papers)}, pages 383--394, Online. Association for Computational Linguistics.

\bibitem[{Ramamurthy et~al.(2022)Ramamurthy, Ammanabrolu, Brantley, Hessel, Sifa, Bauckhage, Hajishirzi, and Choi}]{Ramamurthy2022IsRL}
Rajkumar Ramamurthy, Prithviraj Ammanabrolu, Kiant{\'e} Brantley, Jack Hessel, Rafet Sifa, Christian Bauckhage, Hannaneh Hajishirzi, and Yejin Choi. 2022.
\newblock \href {https://arxiv.org/abs/2210.01241} {Is reinforcement learning (not) for natural language processing?: Benchmarks, baselines, and building blocks for natural language policy optimization}.

\bibitem[{Ran et~al.(2019)Ran, Lin, Li, Zhou, and Liu}]{ran-etal-2019-numnet}
Qiu Ran, Yankai Lin, Peng Li, Jie Zhou, and Zhiyuan Liu. 2019.
\newblock \href {https://doi.org/10.18653/v1/D19-1251} {{N}um{N}et: Machine reading comprehension with numerical reasoning}.
\newblock In \emph{Proceedings of the 2019 Conference on Empirical Methods in Natural Language Processing and the 9th International Joint Conference on Natural Language Processing (EMNLP-IJCNLP)}, pages 2474--2484, Hong Kong, China. Association for Computational Linguistics.

\bibitem[{Rupprecht et~al.(2018)Rupprecht, Laina, Navab, Hager, and Tombari}]{rupprecht2018guide}
Christian Rupprecht, Iro Laina, Nassir Navab, Gregory~D. Hager, and Federico Tombari. 2018.
\newblock \href {http://arxiv.org/abs/1803.11544} {Guide me: Interacting with deep networks}.
\newblock \emph{CoRR}, abs/1803.11544.

\bibitem[{Saunders et~al.(2022)Saunders, Yeh, Wu, Bills, Ouyang, Ward, and Leike}]{saunders2022selfcritiquing}
William Saunders, Catherine Yeh, Jeff Wu, Steven Bills, Long Ouyang, Jonathan Ward, and Jan Leike. 2022.
\newblock \href {https://doi.org/10.48550/ARXIV.2206.05802} {Self-critiquing models for assisting human evaluators}.

\bibitem[{Scheurer et~al.(2022)Scheurer, Campos, Chan, Chen, Cho, and Perez}]{scheurer2022nlfeedback}
Jérémy Scheurer, Jon~Ander Campos, Jun~Shern Chan, Angelica Chen, Kyunghyun Cho, and Ethan Perez. 2022.
\newblock \href {https://doi.org/10.48550/ARXIV.2204.14146} {Training language models with language feedback}.

\bibitem[{Schulman et~al.(2017)Schulman, Wolski, Dhariwal, Radford, and Klimov}]{Schulman2017ProximalPO}
John Schulman, Filip Wolski, Prafulla Dhariwal, Alec Radford, and Oleg Klimov. 2017.
\newblock Proximal policy optimization algorithms.
\newblock \emph{ArXiv}, abs/1707.06347.

\bibitem[{Sheng et~al.(2020)Sheng, Chang, Natarajan, and Peng}]{sheng-etal-2020-towards}
Emily Sheng, Kai-Wei Chang, Prem Natarajan, and Nanyun Peng. 2020.
\newblock \href {https://doi.org/10.18653/v1/2020.findings-emnlp.291} {Towards {C}ontrollable {B}iases in {L}anguage {G}eneration}.
\newblock In \emph{Findings of the Association for Computational Linguistics: EMNLP 2020}, pages 3239--3254, Online. Association for Computational Linguistics.

\bibitem[{Shinn et~al.(2023)Shinn, Cassano, Berman, Gopinath, Narasimhan, and Yao}]{shinn2023reflexion}
Noah Shinn, Federico Cassano, Edward Berman, Ashwin Gopinath, Karthik Narasimhan, and Shunyu Yao. 2023.
\newblock \href {http://arxiv.org/abs/2303.11366} {Reflexion: Language agents with verbal reinforcement learning}.

\bibitem[{Shwartz et~al.(2020)Shwartz, West, Le~Bras, Bhagavatula, and Choi}]{shwartz-etal-2020-unsupervised}
Vered Shwartz, Peter West, Ronan Le~Bras, Chandra Bhagavatula, and Yejin Choi. 2020.
\newblock \href {https://doi.org/10.18653/v1/2020.emnlp-main.373} {Unsupervised commonsense question answering with self-talk}.
\newblock In \emph{Proceedings of the 2020 Conference on Empirical Methods in Natural Language Processing (EMNLP)}, pages 4615--4629, Online. Association for Computational Linguistics.

\bibitem[{Stiennon et~al.(2020)Stiennon, Ouyang, Wu, Ziegler, Lowe, Voss, Radford, Amodei, and Christiano}]{10.5555/3495724.3495977}
Nisan Stiennon, Long Ouyang, Jeff Wu, Daniel~M. Ziegler, Ryan Lowe, Chelsea Voss, Alec Radford, Dario Amodei, and Paul Christiano. 2020.
\newblock Learning to summarize from human feedback.
\newblock In \emph{Proceedings of the 34th International Conference on Neural Information Processing Systems}, NIPS'20, Red Hook, NY, USA. Curran Associates Inc.

\bibitem[{Talmor et~al.(2020)Talmor, Tafjord, Clark, Goldberg, and Berant}]{NEURIPS2020_e992111e}
Alon Talmor, Oyvind Tafjord, Peter Clark, Yoav Goldberg, and Jonathan Berant. 2020.
\newblock \href {https://proceedings.neurips.cc/paper/2020/file/e992111e4ab9985366e806733383bd8c-Paper.pdf} {Leap-of-thought: Teaching pre-trained models to systematically reason over implicit knowledge}.
\newblock In \emph{Advances in Neural Information Processing Systems}, volume~33, pages 20227--20237. Curran Associates, Inc.

\bibitem[{Tandon et~al.(2022)Tandon, Madaan, Clark, and Yang}]{tandon-etal-2022-learning}
Niket Tandon, Aman Madaan, Peter Clark, and Yiming Yang. 2022.
\newblock \href {https://doi.org/10.18653/v1/2022.findings-naacl.26} {Learning to repair: Repairing model output errors after deployment using a dynamic memory of feedback}.
\newblock In \emph{Findings of the Association for Computational Linguistics: NAACL 2022}, pages 339--352, Seattle, United States. Association for Computational Linguistics.

\bibitem[{Wang et~al.(2022)Wang, Wei, Schuurmans, Le, Chi, Narang, Chowdhery, and Zhou}]{wang2022consistency}
Xuezhi Wang, Jason Wei, Dale Schuurmans, Quoc Le, Ed~Chi, Sharan Narang, Aakanksha Chowdhery, and Denny Zhou. 2022.
\newblock \href {https://doi.org/10.48550/ARXIV.2203.11171} {Self-consistency improves chain of thought reasoning in language models}.

\bibitem[{Wang et~al.(2023)Wang, Wei, Schuurmans, Le, Chi, Narang, Chowdhery, and Zhou}]{wang2023selfconsistency}
Xuezhi Wang, Jason Wei, Dale Schuurmans, Quoc~V Le, Ed~H. Chi, Sharan Narang, Aakanksha Chowdhery, and Denny Zhou. 2023.
\newblock \href {https://openreview.net/forum?id=1PL1NIMMrw} {Self-consistency improves chain of thought reasoning in language models}.
\newblock In \emph{The Eleventh International Conference on Learning Representations}.

\bibitem[{Ward et~al.(2022)Ward, Belardinelli, and Toni}]{Ward2022ArgumentativeRL}
Francis~Rhys Ward, Francesco Belardinelli, and Francesca Toni. 2022.
\newblock Argumentative reward learning: Reasoning about human preferences.
\newblock \emph{Workshop on Human-Machine Collaboration and Teaming at ICML}, abs/2209.14010.

\bibitem[{Wei et~al.(2022)Wei, Wang, Schuurmans, Bosma, Chi, Le, and Zhou}]{wei2022chain}
Jason Wei, Xuezhi Wang, Dale Schuurmans, Maarten Bosma, Ed~H. Chi, Quoc Le, and Denny Zhou. 2022.
\newblock \href {http://arxiv.org/abs/2201.11903} {Chain of thought prompting elicits reasoning in large language models}.
\newblock \emph{CoRR}, abs/2201.11903.

\bibitem[{Weidinger et~al.(2021)Weidinger, Mellor, Rauh, Griffin, Uesato, Huang, Cheng, Glaese, Balle, Kasirzadeh, Kenton, Brown, Hawkins, Stepleton, Biles, Birhane, Haas, Rimell, Hendricks, Isaac, Legassick, Irving, and Gabriel}]{ethical-social-risks}
Laura Weidinger, John Mellor, Maribeth Rauh, Conor Griffin, Jonathan Uesato, Po{-}Sen Huang, Myra Cheng, Mia Glaese, Borja Balle, Atoosa Kasirzadeh, Zac Kenton, Sasha Brown, Will Hawkins, Tom Stepleton, Courtney Biles, Abeba Birhane, Julia Haas, Laura Rimell, Lisa~Anne Hendricks, William~S. Isaac, Sean Legassick, Geoffrey Irving, and Iason Gabriel. 2021.
\newblock \href {http://arxiv.org/abs/2112.04359} {Ethical and social risks of harm from language models}.
\newblock \emph{CoRR}, abs/2112.04359.

\bibitem[{Welleck et~al.(2022)Welleck, Lu, West, Brahman, Shen, Khashabi, and Choi}]{welleck2022generating}
Sean Welleck, Ximing Lu, Peter West, Faeze Brahman, Tianxiao Shen, Daniel Khashabi, and Yejin Choi. 2022.
\newblock \href {https://doi.org/10.48550/ARXIV.2211.00053} {Generating sequences by learning to self-correct}.

\bibitem[{Welleck et~al.(2023)Welleck, Lu, West, Brahman, Shen, Khashabi, and Choi}]{welleck2023generating}
Sean Welleck, Ximing Lu, Peter West, Faeze Brahman, Tianxiao Shen, Daniel Khashabi, and Yejin Choi. 2023.
\newblock \href {https://openreview.net/forum?id=hH36JeQZDaO} {Generating sequences by learning to self-correct}.
\newblock In \emph{The Eleventh International Conference on Learning Representations}.

\bibitem[{Weston(2016)}]{weston2016dialog}
Jason Weston. 2016.
\newblock \href {https://doi.org/10.48550/ARXIV.1604.06045} {Dialog-based language learning}.

\bibitem[{Wolf et~al.(2020)Wolf, Debut, Sanh, Chaumond, Delangue, Moi, Cistac, Rault, Louf, Funtowicz, Davison, Shleifer, von Platen, Ma, Jernite, Plu, Xu, Le~Scao, Gugger, Drame, Lhoest, and Rush}]{wolf-etal-2020-transformers}
Thomas Wolf, Lysandre Debut, Victor Sanh, Julien Chaumond, Clement Delangue, Anthony Moi, Pierric Cistac, Tim Rault, Remi Louf, Morgan Funtowicz, Joe Davison, Sam Shleifer, Patrick von Platen, Clara Ma, Yacine Jernite, Julien Plu, Canwen Xu, Teven Le~Scao, Sylvain Gugger, Mariama Drame, Quentin Lhoest, and Alexander Rush. 2020.
\newblock \href {https://doi.org/10.18653/v1/2020.emnlp-demos.6} {Transformers: State-of-the-art natural language processing}.
\newblock In \emph{Proceedings of the 2020 Conference on Empirical Methods in Natural Language Processing: System Demonstrations}, pages 38--45, Online. Association for Computational Linguistics.

\bibitem[{Wu et~al.(2018)Wu, Tian, Qin, Lai, and Liu}]{wu-etal-2018-study}
Lijun Wu, Fei Tian, Tao Qin, Jianhuang Lai, and Tie-Yan Liu. 2018.
\newblock \href {https://doi.org/10.18653/v1/D18-1397} {A study of reinforcement learning for neural machine translation}.
\newblock In \emph{Proceedings of the 2018 Conference on Empirical Methods in Natural Language Processing}, pages 3612--3621, Brussels, Belgium. Association for Computational Linguistics.

\bibitem[{Xie and Sun(2019)}]{Xie2019AGT}
Zhipeng Xie and Shichao Sun. 2019.
\newblock A goal-driven tree-structured neural model for math word problems.
\newblock In \emph{International Joint Conference on Artificial Intelligence}.

\bibitem[{Yao et~al.(2023)Yao, Zhao, Yu, Du, Shafran, Narasimhan, and Cao}]{yao2023react}
Shunyu Yao, Jeffrey Zhao, Dian Yu, Nan Du, Izhak Shafran, Karthik~R Narasimhan, and Yuan Cao. 2023.
\newblock \href {https://openreview.net/forum?id=WE_vluYUL-X} {React: Synergizing reasoning and acting in language models}.
\newblock In \emph{The Eleventh International Conference on Learning Representations}.

\bibitem[{Ye and Durrett(2022)}]{ye2022the}
Xi~Ye and Greg Durrett. 2022.
\newblock \href {https://openreview.net/forum?id=Bct2f8fRd8S} {The unreliability of explanations in few-shot prompting for textual reasoning}.
\newblock In \emph{Advances in Neural Information Processing Systems}.

\bibitem[{Zhang et~al.(2020)Zhang, Wang, Lee, Bin, Wang, Shao, and Lim}]{zhang-etal-2020-graph-tree}
Jipeng Zhang, Lei Wang, Roy Ka-Wei Lee, Yi~Bin, Yan Wang, Jie Shao, and Ee-Peng Lim. 2020.
\newblock \href {https://doi.org/10.18653/v1/2020.acl-main.362} {Graph-to-tree learning for solving math word problems}.
\newblock In \emph{Proceedings of the 58th Annual Meeting of the Association for Computational Linguistics}, pages 3928--3937, Online. Association for Computational Linguistics.

\bibitem[{Ziegler et~al.(2019)Ziegler, Stiennon, Wu, Brown, Radford, Amodei, Christiano, and Irving}]{ziegler2019finetuning}
Daniel~M. Ziegler, Nisan Stiennon, Jeffrey Wu, Tom~B. Brown, Alec Radford, Dario Amodei, Paul Christiano, and Geoffrey Irving. 2019.
\newblock \href {https://doi.org/10.48550/ARXIV.1909.08593} {Fine-tuning language models from human preferences}.

\end{thebibliography}

\clearpage

\appendix

\section{Additional Results} 

\subsection{More details about the quality of the feedback} Please note we also include instances where the critic feedback should say the solution is correct and hence generate 'No'. Our exact match metric is not order-sensitive. We extract the sentences and match them individually to the oracle answers. Since we focused only on the semi-structured critic feedback, automatic evaluation can already capture (measure effectively) the quality of the feedback. 

\subsection{Details about ReACT and Self-consistency and Self-Correct} \label{sec:react}
The ReACT method consists of the reason model (Reason-Only) LLM  (GPT-3.5), which generates a single thought at each step, and the Action model LLM (another GPT-3.5) does the calculation and generates the intermediate outputs (observations). We propose to refine the intermediate steps generated by the above steps and report the results below. Please note ReAct is approx \textbf{3-4} times more expensive than GPT-3.5 + CoT. In our experiments, we assumed $3$ reasoning steps for ReACT and a sample size of $5$ for self-consistency to be more cost-effective. Interestingly, we observe that ReACT perform similarly to CoT for the SVAMP dataset. One intuitive reason is that the SVAMP dataset contains questions which require one or two-hop reasoning only. We find that REFINER performs (+2.2) better than Self-correct \citep{welleck2023generating} on the GSM8K dataset, indicating the importance of correcting the intermediate steps can lead to better performance. Please note that we have used GPT-Neo as the generator model and the Unified QA T5-base model as the critic model, consistent with the Self-correct paper by Welleck et al. (2022).

\subsection{More results on SVAMP dataset}
In the MWP, for the answer prediction task, we compare \ourmodel with the previously reported baselines from \citet{jie-etal-2022-learning} including Graph2Tree \citep{zhang-etal-2020-graph-tree} that uses quantity relations using GCN; GTS \citep{Xie2019AGT} which is a sequence-to-tree model that mainly uses a tree-based decoder with GRU; and DeductReasoner \citep{jie-etal-2022-learning} which uses bottom-up DAG-structured decoding. Results of this comparison can be found in Table \ref{tab:results_extra}. For the sNLR task, we also experiment with a critic model trained on 50\% of its original training data and we still observe a performance improvement over the baseline as can be seen in Table \ref{tab:50_snr_results}.

\begin{table}[t!]
\centering
\scalebox{0.7}
{
\begin{tabular}{@{}l@{~~}c@{~~~~}}
\toprule
{\bf Model} & {\bf Accuracy}
\\\midrule
GPT-Neo (1.3B)	& 8.5 \\
GPT-Neo + Self-Correct	& 21.2 \\
GPT-Neo + REFINER	& 23.4 +/- 0.3 \\
\bottomrule
\end{tabular}}
\caption{\textbf{Comparing} REFINER with self-correct on GSM8K dataset}
%\vspace{-1.5em}
\label{tab:trained_critic}
\end{table}

\begin{table}[h]
\centering
\scalebox{0.7}{
\begin{tabular}{@{}l@{~~~~}c@{~}}
\toprule
%{\bf Model} & {\bf EM} & {\bf Acc}\\\midrule
\textbf{Answer Prediction ($y$)}& \textbf{Acc \%}\\
\hline
GTS  & {30.8} \\
Graph2Tree  &{36.5} \\
BERT-Tree & {32.4}\\
Roberta-large-GTS & {41.0} \\ 
Roberta-large-Graph2Tree & {43.8} \\ 
Roberta-large-DeductReasoner & {45.0} \\
Few-Shot GPT-3 & {63.05} \\
Few-Shot GPT-3 + COT & {63.5} \\ 
Few-Shot GPT-3 + COT + \model & {\textbf{66.4}} \\
\bottomrule
\end{tabular}}
\caption{Results on SVAMP dataset}
\label{tab:results_extra}
\end{table}

\section{\ourmodel Framework}
\label{sec:appendix_framework}
\Algref{alg:training_refiner} and \Algref{alg:inference_refiner} outline the training and inference algorithms for \ourmodel. We train a supervised \textsc{critic} model ($\pi_{\beta}$) with the context ($x$) and (plausible or implausible) hypothesis ($z$ or $z'$) as input and the textual feedback as output. Given a context $x$ the generator model ($\pi_{\theta}$) is trained to generate plausible hypotheses.

\begin{algorithm}\caption{\ourmodel Training}
\begin{algorithmic}[1]
%\State Initialize $m \gets$ empty list
\For {E epochs}
    \For {$i (batch) \gets 1$ to $N$}
        %\State Initialize (reward) $r_i \gets 0$, $p_i \gets 1$
        \State Initialize (feedback) $f_0 \gets No$
        \For {$t \gets 1$ to $T$}
            \State {$\hat{z}_{i,t}^{k}$ $\sim$  $\pi_{\theta}(y_i|c_i, f_{t-1}, \hat{z}_{i,t-1})$}
            \State {$f_{t}, \hat{z} \gets \pi_{\beta}(c_i, z_i, \hat{z}_{i,t}^{k})$}
            \State {$\mathcal{L}^{lm}_{i}$} += $-\log p(z_i|c_i, f_{t-1}, \hat{z}_{i,t-1})$

        \EndFor
    \EndFor
\EndFor
\State \Return {$\pi_{\theta}$}
\end{algorithmic}
\label{alg:training_refiner}
\end{algorithm}

\begin{algorithm}\caption{REFINER Inference}
\begin{algorithmic}[1]
\State Initialize $answers  \gets$ empty list
\For {$i (batch) \gets 1$ to $N$}
    \State Initialize (reward) $r_i \gets 0$, $p_i \gets 1$
    \State Initialize (hint) $h_0, \hat{y}_{i,0} \gets No, []$
    \For {(turn) $t \gets 1$ to $T$}
        \State {$\hat{y}$ $\gets$  $\pi_{\theta}(y_i|c_i, h_{t-1}, \hat{y}_{i,t-1})$}
        \State {$h_{t} \gets \pi_{\beta}(c_i,  \hat{y}_{i})$}
            
        \If {$h_{t}$ == No}
                \State $answers$.append($\hat{y}$)
                \State break
        \EndIf
    \EndFor
    \State $answers$.append($\hat{y}$)
\EndFor
\State \Return $answers$
\end{algorithmic}
\label{alg:inference_refiner}
\end{algorithm}

\section{Datasets and Models}
In Table \ref{tab:data_stat} and Table \ref{tab:dataset_details}, we report the data statistics and dataset details. In Table \ref{tab:model_stat}, we report the details of the used models. Our research is conducted solely on datasets that are in the English language. 

\begin{table}[h]
\small
\centering
{
\begin{tabular}{@{}lccc@{}}
\toprule
{\bf Task} & {\bf Train}& {\bf Dev} & {\bf Test}\\
\midrule
MWP & {3,138}& {--} & {1000} \\
sNLR& {1000}& {5000} & {5000} \\
MS & {10000}& {1000} & {1000}\\
GSM8k & {--} & {--}& 1319 \\ 
%Counterfactual Example (NLG) & {47763}& {1532} & {3059} \\
\bottomrule
\end{tabular}}
\caption{Dataset Statistics:
nb.\ of instances.}\label{tab:data_stat}
\end{table} 

\begin{table}[h]
\small
\centering
{
\begin{tabular}{@{}lc@{}}
\toprule
{\bf Model} & {\bf Parameter Size}\\
\midrule
UQA-base & {220M}\\
REFINER$_{base}$ & {440M}\\
UQA-large& {770M}\\
REFINER$_{large}$ & {990M}\\
GPT3.5 & {175B}\\
%Counterfactual Example (NLG) & {47763}& {1532} & {3059} \\
\bottomrule
\end{tabular}}
\caption{Model Sizes.}\label{tab:model_stat}
\end{table}

\begin{table*}[h]
\small
\centering
{
\begin{tabular}{@{}lll@{~~~}l@{}}
\toprule
{\bf Dataset/Tools} & {\bf Citation}& {\bf Link} & \bf{License}\\
\midrule
SVAMP & \citet{patel-etal-2021-nlp} & \url{https://github.com/arkilpatel/SVAMP} & {MIT License}\\
GSM8k & \citet{cobbe2021gsm8k} & \url{https://github.com/openai/grade-school-math} & {MIT License}\\
sNLR & \citet{liang2022holistic} & \url{https://github.com/stanford-crfm/helm} & Apache License\\
Moral Norm & \citet{emelin-etal-2021-moral} & \url{https://github.com/demelin/moral_stories} & {MIT License}\\
HuggingFace  & \citet{wolf-etal-2020-transformers} & \url{https://github.com/huggingface/transformers} & {Apache License}\\
\bottomrule
\end{tabular}}
\caption{More details about datasets and Tools}\label{tab:dataset_details}
\end{table*}

\begin{figure*}[t]
  \centering
    \includegraphics[height=5cm, width=0.75\paperwidth]{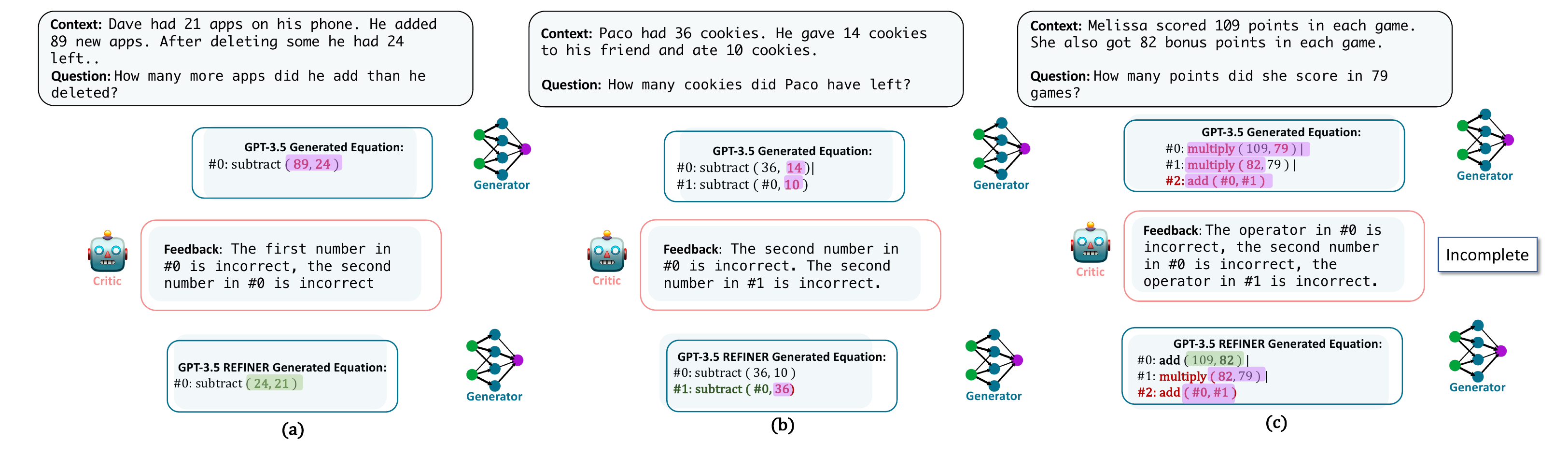}
    \caption{\textbf{Examples.} REFINER on MWP task. %Equation generated by \textsc{generator} model (GPT3.5), \textit{feedback} generated by the \textsc{critic} model. 
    There are different scenarios are highlighted in the figure, where (a) the \textsc{critic} model provides correct feedback, \textsc{generator} model utilizes the feedback and fixes the incorrect equation, (b) the \textsc{critic} model provides a \textit{correct} feedback however, \textsc{generator} model fails to fix the \textit{incorrect} equation, and (c) the \textsc{critic} model provides an \textit{incomplete} feedback \textsc{generator} model partially fixes the incorrect equation.}
    \label{fig:qualitative_analysis}
\end{figure*}

\begin{table}[t!]
\small
\centering
\begin{tabular}{@{}l@{~~~~}c@{~~~}c@{~~~}}
\toprule
{\bf Model} & {\bf Eq. ($z$)} & {\bf Ans. (${y}$)}\\\midrule

UQA-large & {\cellcolor{NextBlue}46.7} & {\color{gray} --} \\
UQA-large + PPO & \cellcolor{NextBlue}{48.2} & {\color{gray} --} \\
\model$_{large}$ & \cellcolor{NextBlue}\textbf{53.8} & {\color{gray} --} \\
\model$_{large}$ + {\color{orange} Oracle} (T=3)& \cellcolor{NextBlue}{68.1} & {\color{gray} --} \\
\midrule
GPT-$3.5$ + CoT & {59.3} & \cellcolor{Nextblush}{63.5} \\
%GPT3.5 + REFINER$_{critic}$ & \cellcolor{NextBlue}\textbf{59.6} & {--} \\
GPT-$3.5$ + CoT + REFINER$_{critic}$ & {62.3} & \cellcolor{Nextblush}\textbf{66.4} \\
GPT-$3.5^\star$ + CoT & {64.1} & \cellcolor{Nextblush}{67.1} \\
%GPT3.5 + REFINER$_{critic}$ & \cellcolor{NextBlue}\textbf{59.6} & {--} \\
GPT-$3.5^\star$ + CoT + REFINER$_{critic}$ & \textbf{67.3} & \cellcolor{Nextblush}\textbf{70.6} \\

\bottomrule
\end{tabular}
\caption{Results on MWP. Eq.: Equation, Ans. Answer. Comparison of \ourmodel with baselines on the SVAMP dataset. GPT-$3.5$: code-DaVinci-002, GPT-$3.5^\star$: text-DaVinci-002 For models other than GPT3.5, the answer can be obtained via symbolic execution of the equation and is thus a function of the validity of the equation. For GPT3.5, the model is few-shot prompted to either generate the equation with variable names $z$, or generate the answer $y$.
}
\label{tab:results_mwp_app}
\end{table}

\if false
\begin{table}[t!]
\centering
{
\begin{tabular}{@{}l@{~~}c@{~~~~} c@{~~~~}}
\toprule
{\bf Model} & {\bf EM} & {\bf Acc}\\\midrule
\rowcolor{gray!25}
\textbf{SVAMP} - Equation Generation  &  &\\

\model$_{base}$ - critic$_{inference}$ & {39.8} & --\\
\model$_{base}$ - critic$_{inference}$ - exp & {37.4} & --\\

\model$_{base}$ - critic$_{training}$ & {34.1} & --\\ 
%TAC$_{base}$ + oracle critic & {} & \\
\model$_{large}$ - critic$_{inference}$ & {48.44} & --\\
\model$_{large}$ - critic$_{inference}$ - exp & {45.55} & --\\
%TAC$_{large}$ + oracle critic & {} & \\
\rowcolor{gray!25}
\textbf{SNR} &  &\\
\model$_{base}$ - critic$_{inference}$ & {92.92} & {97.32}\\
\model$_{base}$ - critic$_{inference}$ - exp & {} & {}\\
\rowcolor{gray!25}
\textbf{Moral Norm} &  &\\
\bottomrule
\end{tabular}}
\caption{\textbf{Ablation Result.} exp: Exploration}
\label{tab:results}
\end{table}
\fi
\begin{table}[t!]
\centering
\begin{tabular}{@{}l@{~}c@{~~~~}c@{~~~~}}
\toprule
{\bf Model} & {\bf IR} & {\bf C}\\\midrule
\rowcolor{gray!25}
\textbf{50\% training data} &  &\\
T5-base & {84.28 $\pm$ 0.5} & {88.86}\\
\ourmodel$_{base}$ & {\bf{88.26} $\pm$ 0.8} & \bf{94.26} \\
\model$_{base}$ + Oracle & {91.11 $\pm$ 05} & {97.28}\\
\bottomrule
\end{tabular}
\caption{Results on SNR dataset. IR: Inference Rules, C: Consequent}
\label{tab:50_snr_results}
\end{table}

\if false
\begin{table}[t!]
\centering
\begin{tabular}{@{}l@{~}c@{~~~~}c@{~~~~}}
\toprule
{\bf Model} & {\bf Scenario 1} & {\bf Scenario 2}\\\midrule
\rowcolor{gray!25}
\textbf{Few-Shot Setting} &  &\\
GPT-3 + COT & {20.9} & {17.8} \\
GPT-3 + COT + TAC & \bf{30.3} & \bf{26.8} \\
\bottomrule
\end{tabular}
\caption{Analysis on SNR dataset. IR: Inference Rules, C: Consequent}
\label{tab:results}
\end{table}
\fi

\begin{figure*}[t]
\small
  \centering
    \includegraphics[scale=0.1, width=0.65\paperwidth]{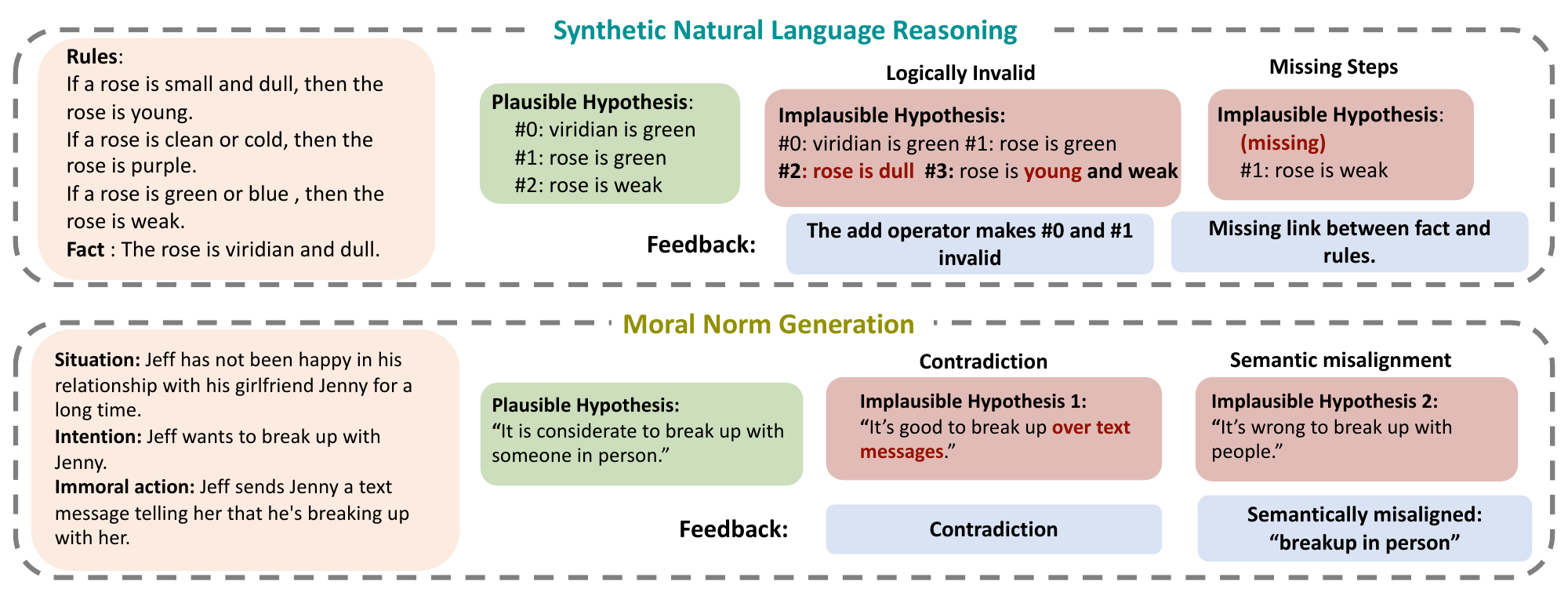}
    \caption{\textbf{Feedback Data Generation}. The top row illustrates an example from the sNLR task, where the error types are \textit{logically invalid}, \textit{missing links}, and \textit{missing implicit knowledge steps}. The bottom row illustrates an example from moral norm generation, where the error types are \textit{contradiction} and \textit{semantic misalignment}. We perturbed used the plausible intermediate steps to implausible.}
    \label{fig:error_overview}
    \vspace{-1.5em}
\end{figure*}

\section{Training Details} \label{sec:training_details}
\paragraph{Training Details.} For each task, we train a UnifiedQa-\textsc{T5}-base model (UQA-base) \citep{khashabi-etal-2020-unifiedqa} as a critic (\S \ref{sec_3.1:critic_model}). Further evaluation details are provided in Appendix \ref{sec:Appendix_E}.  For exploration (\S \ref{sec:3.2_generator}), we use nucleus sampling with $p = 0.5$. We select the hyper-parameters by the validation loss: for both the generator and critic model, we use the Adam optimizer with a learning rate of $1e^{-4}$. Each model is trained for $20$ epochs with early stopping based on validation loss. We trained all models on one A100 GPU. We run our models with $3$ random seeds and report the average results. We perform a binomial sign test. We find that p-values are always <0.05 when we compare REFINER with all the baselines (GPT-3.5, Self-refine, Self-reflection), suggesting our results are not random and significant. For the human study, we selected outputs from the best models (baselines and our model) according to automatic metrics. We train models with $T=3$ iterations. We trained the critic model for 8 hours and trained the generator model for 12 hours. 

At inference time, we use greedy decoding for the generator and critic model with $T=1$ for the automatic critic and $T=3$ for the oracle critic. We evaluate our methods using the metrics presented in the original papers that proposed the tasks. On the MWP and sNLR tasks, we use the exact match (EM) metric for intermediate steps (equation generation and inference rules) and accuracy (Acc) for the final answers. For MS, we conduct a manual evaluation study to assess the relevance of norms and moral actions.\footnote{Since the automatic scores such as BLUE, ROUGE, etc. only account for word level similarity between gold norms or actions and generate norms or actions.}
\section{Qualitative Examples}
\begin{figure*}[t]
  \centering
    \includegraphics[scale=1,height=9cm, width=0.72\paperwidth]{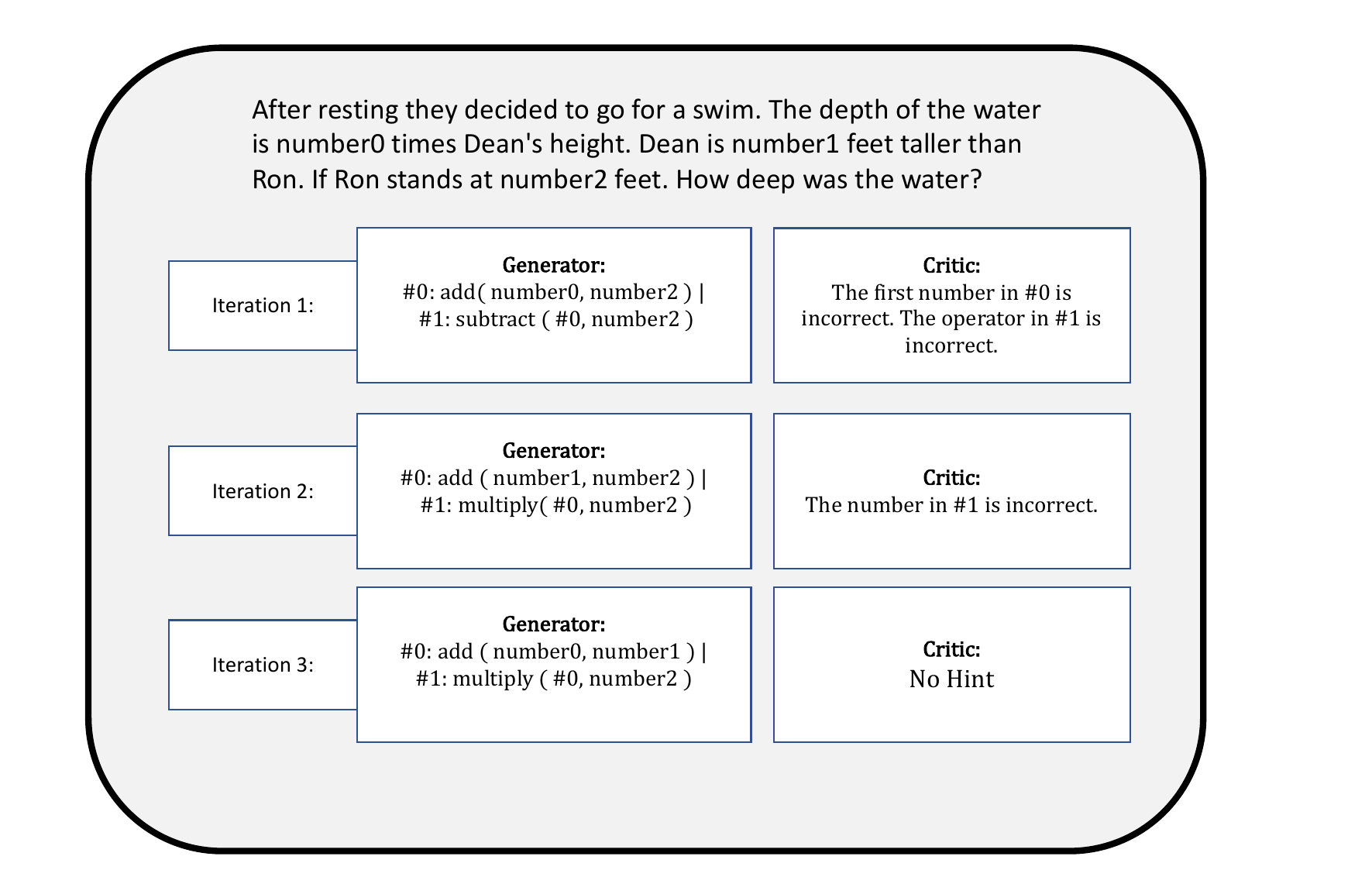}
    \caption{\ourmodel on MWP. The generator's output improves step-wise.}
    \label{fig:example_multistep}
\end{figure*}

Figure \ref{fig:example_multistep} and \ref{table:moral-stories-gen} depict a qualitative example of REFINER where \ourmodel could correct incorrect equations through structured feedback, fixing the operators within a multistep solution. Table \ref{table:moral-stories-gen} shows some qualitatively improved examples for MS. 
\if False
\begin{figure*}[t]
  \centering
    \includegraphics[scale=1,height=9cm, width=0.72\paperwidth]{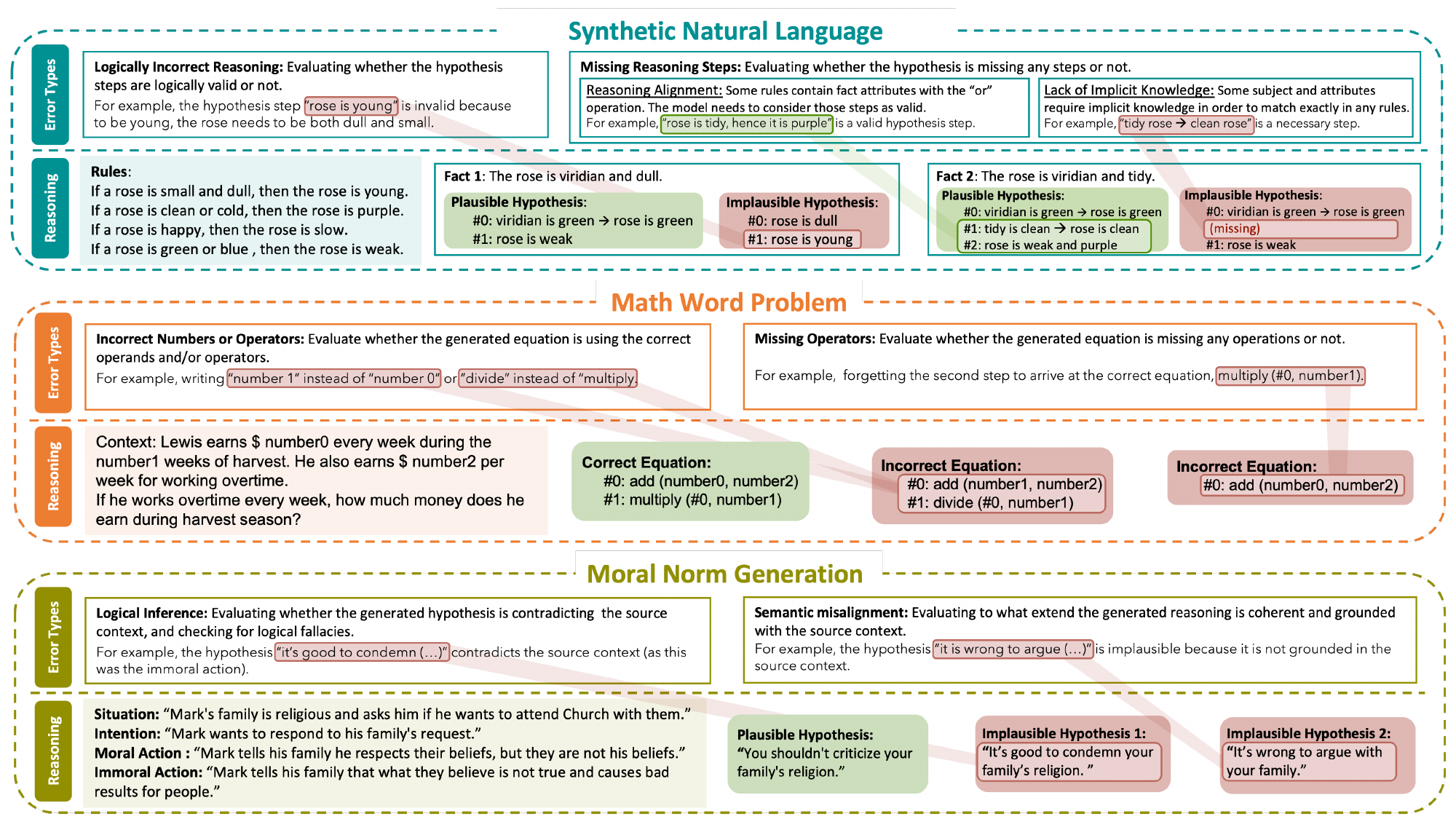}
    \caption{An overview of the three tasks tackled in this paper, with examples of both valid and invalid intermediate reasoning steps, as well as their corresponding fine-grained error types. Notice the \textbf{Missing Steps} error type, in the second task, actually encompasses two error types: reasoning misalignment, derived from not considering the \texttt{or} operation, and lack of implicit knowledge, where implicit knowledge is needed to match the existing rules.}
    \label{fig:error_overview_all}
\end{figure*}
\fi 

\begin{table*}[t!]
\small
\scalebox{0.8}{
\centering
\begin{tabular}{@{}l@{~}l@{~~~}l@{~~}l@{~~~~}}
\toprule
{\bf Task} & {\bf Error Types} & \bf{Structured Feedback} &{\bf Human Readable Feedback}\\\midrule
\textbf{MWP} & {Incorrect Numbers}  & {$\langle error type, position, equation-number \rangle$} & The \texttt{position}  number in \texttt{equation-number} is incorrect. \\
 & {Incorrect Operators} & {$\langle error type, equation-number \rangle$} & The operator in \texttt{equation-number} is incorrect. \\
 & {Missing Operators} & {$\langle error type \rangle$} & An operator is missing. \\
\textbf{sNLR} & {Logically Incorrect}  & {$\langle$ X operator, inference rule number $\rangle$} & The \texttt{X operator}  makes \texttt{inference rule number} invalid. \\
 & {Missing Lookup Step} & {$\langle error type \rangle$} & Missing link between the fact and the rules. \\
 & {Missing Implicit Knowledge Step} & {$\langle error type \rangle$} & The implicit knowledge is missing. \\
\bottomrule
\end{tabular}}
\caption{Feedback Templates}
\label{tab:templates}
\end{table*}

\section{Feedback Data Generation}\label{sec:feedback_gen}

\subsection{Rule-based Perturbation}\label{sec:perturbation_gen}
Based on these error types, we perturb the plausible hypotheses ($z$) in the training data and collect a pool of data $D$ ($x$: input, $z$: plausible hypothesis, $z'$: implausible hypothesis). We perturb by omitting, replacing or adding some tokens or some rules from the plausible hypothesis to automatically create an implausible hypothesis. For example, in \Figref{fig:error_overview}, for sNLR we omit a few inference steps from the correct hypothesis "\texttt{\#0: viridian is green, \#1: rose is green}" and create an incorrect (incomplete) hypothesis (see \Figref{fig:error_overview}). Since our perturbations are based on logic and reasoning errors, we create structured feedback $f$ for every example ($x, z, z'$) by stating the error type that occurs in $z'$ but not in $z$ (see Table \ref{tab:define_error_feedbacks}). The basic structure of feedback $f$ for these tasks is $\langle$\textit{error type, position (optional), hint (optional)}$\rangle$, where position denotes the error position in the implausible hypothesis (see Appx Table \ref{tab:define_error_feedbacks}). For example, in the previous scenario, we create feedback ``\textit{Missing link between fact and rules}''. Despite the simplicity of the strategy we used for our tasks, this approach is easily generalisable to other reasoning tasks. 

For MWP and sNLR problems, the underlying reasoning requires symbolic systems with closed-world rules. Hence, we consider a simple rule-based method to automatically generate the pairs of errors and their corresponding structured feedback by considering the error types and position of the errors (see \Figref{fig:error_overview} and Table \ref{tab:define_error_feedbacks}). 

In the moral norm generation task, we consider two kinds of fine-grained errors: \textit{logical contradiction} and \textit{semantic misalignment} (incoherent, uninformative). Moral norms are people’s subjective judgments about the character and actions mentioned in the context. Each moral norm is a combination of two components (implicit structure): a moral judgment \texttt{[You shouldn't]} and an action \texttt{[criticize your family's religion]}. 
Firstly, to create \textit{logical contradictions}, we use the concept of deontic logic from \citet{kiehne-emnlp-2022} and derive new norms contrary to those of Moral Stories. Hence, we replace the correct moral judgments in the plausible hypothesis with inverse judgments. For example, replacing \texttt{[You shouldn't]} from the plausible hypothesis to \texttt{[It's good]}, as depicted in \Figref{fig:error_overview}. To scale such inverse norms (\textit{implausible hypothesis}), we paraphrase them by substituting the adjectives with synonyms from WordNet. 
Secondly, to create \textit{semantic misalignments}, we must collect implausible hypotheses that are either misaligned with the plausible hypothesis or incomplete in nature. 
To create them, we replace the correct action (verb phrase) from the plausible hypothesis with random verb phrases selected from the context of the plausible hypothesis.

\subsection{Synthetic Feedback Generation}\label{sec:synthetic_gen} 

We used a few-shot setting where we varied the
instruction, the number of demonstrations, and the formatting of the demonstrations. Since data generation with GPT-$3.5$ is expensive, we generated $30$K, $20$K, and $30$K implausible hypotheses for MWP, sNLR and MS tasks, respectively.

\begin{table}
\centering
\resizebox{\linewidth}{!}{
\footnotesize
\centering
\begin{tabular}{l}
\toprule
\textbf{Initial PROMPT: Math Word Problem} \\
\midrule
\texttt{You are a helpful assistant for math word problems.} \\
\texttt{We will provide you with a math word problem,} \\
\texttt{and your task is to generate the intermediate mathematical } \\ \texttt{equations as a step for solving the problem} \\
\texttt{and the final correct answer.  Here are two examples: } \\ \\
\texttt{“Question : ” <Problem Statements> Let’s think step by step }\\
\texttt{<equation> Answer: <answer> } \\ \\
\texttt{“Question: ” <Problem Statements> Let’s think step by step }\\
\texttt{<equation> Answer: <answer> } \\ \\
\texttt{“Question: ” <Problem Statements> Let’s think step by step  }\\
\bottomrule
\end{tabular}}
\caption{Prompts used for generating correct answer given a math word problem}
\label{table:event_extraction_prompt}
\end{table}

\begin{table}
\centering
\resizebox{\linewidth}{!}{
\footnotesize
\centering
\begin{tabular}{l}
\toprule
\textbf{REFINEMENT PROMPT: Math Word Problem} \\
\midrule
\texttt{You are a helpful assistant for math word problems.} \\
\texttt{We will provide you with a math word problem } \\
\texttt{a solution (containing an equation and an answer), } \\ \texttt{and feedback on the solution.} \\
\texttt{Your task is to generate a refined intermediate equation} \\
\texttt{as a step and the final correct answer. } \\ 
\texttt{Here are two examples:} \\ \\ 
\texttt{“Question : ” <Problem Statements> Let’s think step by step }\\
\texttt{<equation> Answer: <answer> } \\ 
\texttt{Feedback: <feedback> <equation> Answer: <answer>} \\ \\
\texttt{“Question: ” <Problem Statements> Let’s think step by step }\\
\texttt{<equation> Answer: <answer> } \\ 
\texttt{Feedback: <feedback> <equation> Answer: <answer>} \\ \\
\texttt{“Question: ” <Problem Statements> Let’s think step by step  }\\
\texttt{<equation> Answer: <answer> } \\
\texttt{Feedback: <feedback> } \\
\bottomrule
\end{tabular}}
\caption{Prompts used for generating correct answer given a math word problem}
\label{table:event_extraction_prompt}
\end{table}

\begin{table}
\centering
\resizebox{\linewidth}{!}{
\footnotesize
\centering
\begin{tabular}{l}
\toprule
\textbf{PROMPT: Synthetic Incorrect Instance Generation} \\
\midrule
\texttt{You are a helpful assistant for} \\ 
\texttt{generating counterfactual reasoning steps.} \\
\texttt{We will provide you with a problem, an error type} \\
\texttt{and a correct intermediate reasoning step.} \\
\texttt{Your task is to generate an incorrect reasoning step } \\
\texttt{based on the error type. } \\
\texttt{Here are a few examples for each error type: } \\ \\
\texttt{“Question : ” <Problem Statements> Let’s think step by step }\\
\texttt{<correct intermediate steps> Error type: <error type> } \\ 
\texttt{Counterfactual: <incorrect intermediate steps> } \\ \\
\texttt{“Question: ” <Problem Statements> Let’s think step by step }\\
\texttt{<correct intermediate steps> Error type: <error type> } \\ 
\texttt{Counterfactual: <incorrect intermediate steps> } \\ \\
\texttt{“Question: ” <Problem Statements> Let’s think step by step  }\\
\texttt{<correct intermediate steps> Error type: <error type> } \\ 
\texttt{Counterfactual: }\\
\bottomrule
\end{tabular}}
\caption{Prompts used for generating synthetic incorrect instances }
\label{table:event_extraction_prompt}
\end{table}

\begin{table*}[h]
\small 
\begin{tabular}{|l|p{0.7\textwidth}|} 
\hline
& \\
\textbf{Situation:} & Jim was in class when his classmate answered one of the teacher's questions wrong. \\ 
\textbf{Intention:} & Jim wants his classmate \yellowbox{to know the answer} was wrong. \\
\textbf{Moral Action:} & Jim tells his classmate the right answer and offers to help him after school. \\
\textbf{Immoral Action:} & Jim starts to laugh at his classmate and tells him he \yellowbox{is stupid} for not knowing the answer. \\
\textbf{Plausible Hypothesis:} & \bluebox{It's hurtful} \graybox{to make fun of your classmates.} \\
\textbf{Contradictory Hypothesis:} & \bluebox{You should} \graybox{make fun of your classmates.} \\
\textbf{Implausible Hypothesis 1:} & \bluebox{It's good} \yellowbox{to know the answer.} \\
\textbf{Implausible Hypothesis 2:} & \bluebox{It's wrong} \yellowbox{to be stupid.} \\
& \\
\hline
& \\
\textbf{Situation:} & Jenny has been going out with a guy for a while, but she feels it isn't working out. \\
\textbf{Intention:} & Jenny wants \yellowbox{to end the relationship with the guy.} \\
\textbf{Moral Action:} & Jenny meets the guy and gently says that she wants to end the relationship.\\
\textbf{Immoral Action:} & Jenny \yellowbox{sends} the guy \yellowbox{a text message} that informs him that the relationship is over. \\
\textbf{Plausible Hypothesis:} & \bluebox{It's good} \graybox{to break up with someone face to face.} \\
\textbf{Contradictory Hypothesis:} & \bluebox{It's bad} \graybox{to break up with someone face to face. } \\
\textbf{Implausible Hypothesis 1:} & \bluebox{You should not} \yellowbox{send a text message} \\
\textbf{Implausible Hypothesis 2:} & \bluebox{It's bad} \yellowbox{to end the relationship with the guy.} \\
& \\
\hline
& \\
\textbf{Situation:} & Paula is leaving home \yellowbox{to do some shopping} and notices that the dog's water bowl is empty. \\
\textbf{Intention:} & Paula wants \yellowbox{to leave the dog at home.} \\
\textbf{Moral Action:} & Paula fills the bowl with water before leaving dog alone in the house.\\
\textbf{Immoral Action:} & Paula decides to fill the bowl when she gets home and leaves the dog with no water. \\
\textbf{Plausible Hypothesis:} & \bluebox{You must always} \graybox{be kind to animals }\\
\textbf{Contradictory Hypothesis:} & \bluebox{It's bad} \graybox{to be kind to animals.} \\
\textbf{Implausible Hypothesis 1:} & \bluebox{It's rude} \yellowbox{to leave the dog at home.}\\
\textbf{Implausible Hypothesis 2:} & \bluebox{It's important} \yellowbox{to do some shopping.} \\
& \\
\hline
\end{tabular}
\caption{Critic training examples from Moral Stories. Norm judgments are highlighted in \bluebox{light blue}, original norm actions are colored in \graybox{light gray} and norm actions extracted from the context are colored in \yellowbox{light yellow}.}
\label{table:moral-stories-ex}
\end{table*}

\begin{table*}[h]
\small 
\begin{tabular}{|l|p{0.7\textwidth}|} 
\hline
& \\
\textbf{Situation:} & Jeff has not been happy in his relationship with his girlfriend Jenny for a long time. \\ 
\textbf{Intention:} & Jeff wants to break up with Jenny. \\
\textbf{Moral Action:} & Jeff goes to Jenny's house and tells her that he wants to break up with her. \\
\textbf{Immoral Action:} & Jeff sends Jenny a text message telling her that he's breaking up with her. \\
\textbf{UQA-T5-large Output:} & \bluebox{It's wrong to break up with someone.} \\
& \greenbox{Jeff calls Jenny and tells her that he's breaking up with her.} \\
& \\
\textbf{\ourmodel Output:} & \bluebox{Its wrong to break up with someone over text.} \\
& \greenbox{Jeff sits down with Jenny and tells her that hes breaking up with her.} \\
& \\
\hline
& \\
\textbf{Situation:} & Jack and his friends are playing in his backyard on a sunny day. \\
\textbf{Intention:} & Jack wants to have fun with his friends. \\
\textbf{Moral Action:} & Jack sets up an inflatable pool and swims in the pool with his friends. \\
\textbf{Immoral Action:} & Jack tosses a ball around with his friends, and he accidentally hit a window of his neighbor. \\
\textbf{UQA-T5-large Output:} & \bluebox{It's wrong to hit someone.} \\
& \greenbox{Jack throws a ball with his friends and hits a window of his neighbor.} \\
& \\
\textbf{\ourmodel Output:} & \bluebox{Its wrong to cause harm to someone elses property.} \\
& \greenbox{Jack tosses a ball around with his friends and hits a tree with it.} \\
& \\
\hline
& \\
\textbf{Situation:} & Jack was invited to a party, but he has the flu. \\
\textbf{Intention:} & Jack wants to attend the party and have fun. \\
\textbf{Moral Action:} & Jack stays home, but Skypes in to the party to socialize.\\
\textbf{Immoral Action:} & Jack goes to the party even though he is sick. \\
\textbf{UQA-T5-large Output:} & \bluebox{It's bad to be sick.} \\
& \greenbox{Jack calls his mom and tells her he is sick.} \\
& \\
\textbf{\ourmodel Output:} & \bluebox{Its bad to spread germs.} \\ & \greenbox{Jack calls his friend and tells him he cant go to the party.} \\
& \\
\hline
\end{tabular}
\caption{\textbf{Moral Stories generations.} Norm outputs (\textit{$z$}) are highlighted in \bluebox{light blue}, moral action outputs (\textit{$y$}) are colored in \greenbox{light green}}.
\label{table:moral-stories-gen}
\end{table*}

\if false 
Example where GPT-3 output were correct for incorrect reasons: 
{
        "actor_input": "<|SIT|> Jake is out with his friends at the club for the night. <|INT|> Jake wants to get drunk. <|I_ACT|> Jake gets super drunk and gets mad at someone for standing next to him he then starts to fight with him and punches him. <|NRM|>",
        "actor_output": "It's mean to pick fights with people at clubs. <|M_ACT|> Jake gets drunk and has fun hanging out with his friends.",
        "generated_moral_action_gpt-3": "Jake gets drunk and has a good time with his friends.",
        "generated_moral_action_gpt-3_critic": "Jake gets drunk and has a good time with his friends.",
        "generated_norm_gpt-3": "It is good to be a good friend.",
        "generated_norm_gpt-3_critic": "It's good to not get into fights.",
        "hint": "not getting into fights."
    },

Example where test our critic was good but GPt-3 didn't picked up! 
 {
        "actor_input": "<|SIT|> Tyrone is starting a fitness routine and is looking for ways to build muscle in a short time. <|INT|> Tyrone wants to do all he can to build muscle. <|I_ACT|> Tyrone starts taking anabolic steroids and eating a lot to promote lots of muscle growth. <|NRM|>",
        "actor_output": "Slow and steady wins the race. <|M_ACT|> Tyrone follows a strict diet and has help from a personal trainer to meet his goals.",
        "generated_moral_action_gpt-3": "Tyrone starts taking protein supplements and eating a lot to promote muscle growth.",
        "generated_moral_action_gpt-3_critic": "Tyrone starts taking protein supplements and eating a lot to promote muscle growth.",
        "generated_norm_gpt-3": "It is good to be healthy.",
        "generated_norm_gpt-3_critic": "It's good to be healthy.",
        "hint": "not taking anabolic steroids."
    },
\fi 
\section{Human Evaluation on Moral Stories} \label{sec:Appendix_E}
\begin{figure*}
    \centering   
    \includegraphics[scale=1, width=0.6\paperwidth]{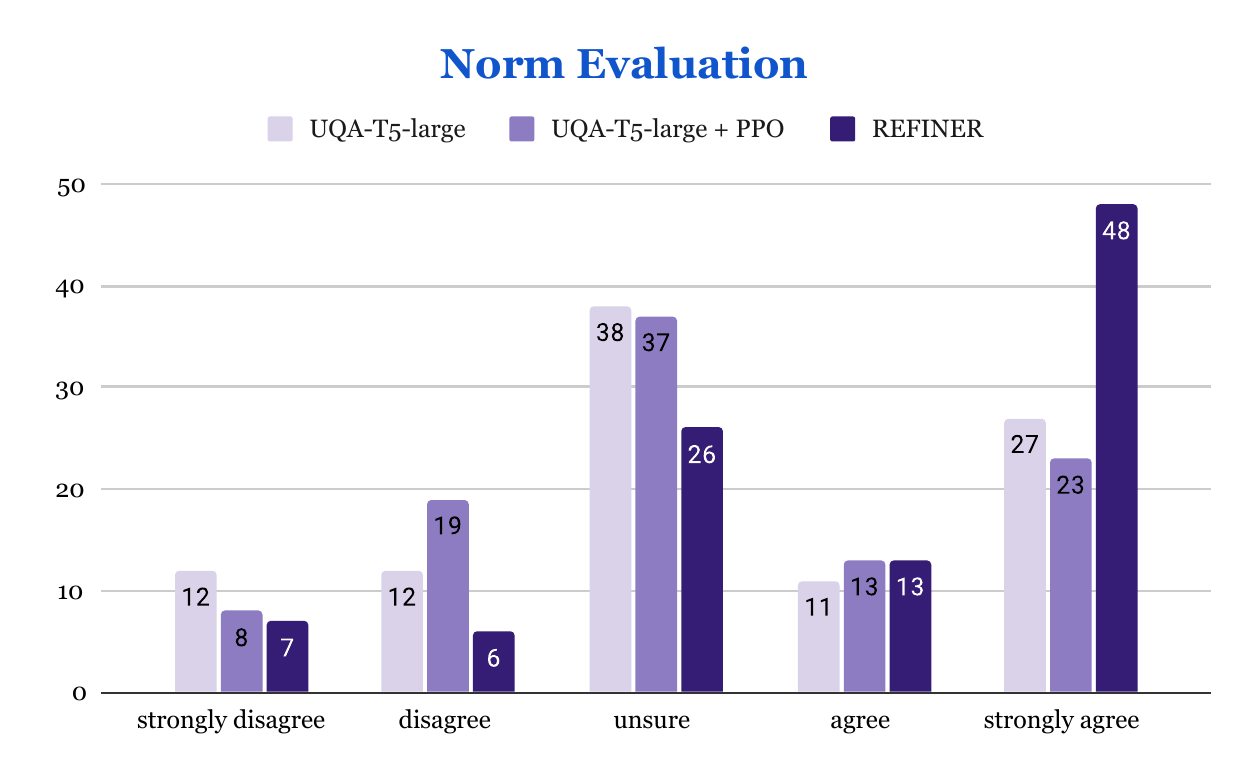}
    \caption{Human Evaluation of Moral Norm on 100 test samples.}
    \label{fig:norm-evaluation}
\end{figure*}

\begin{figure*}
    \centering   
    \includegraphics[scale=1, width=0.6\paperwidth]{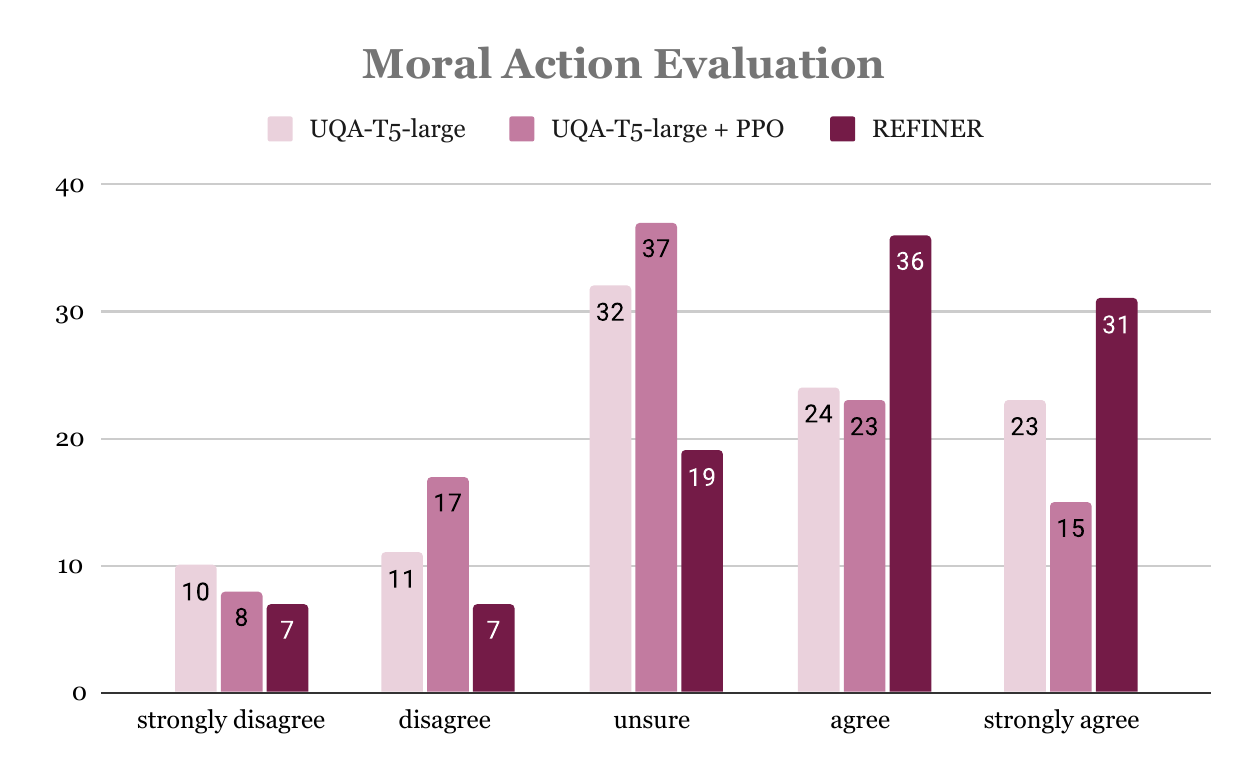}
    \caption{Human Evaluation of Moral Action on 100 test samples.}
    \label{fig:moral-action-evaluation}
\end{figure*}

\begin{figure*}
    \centering   
    \includegraphics[scale=1, width=0.7\paperwidth]{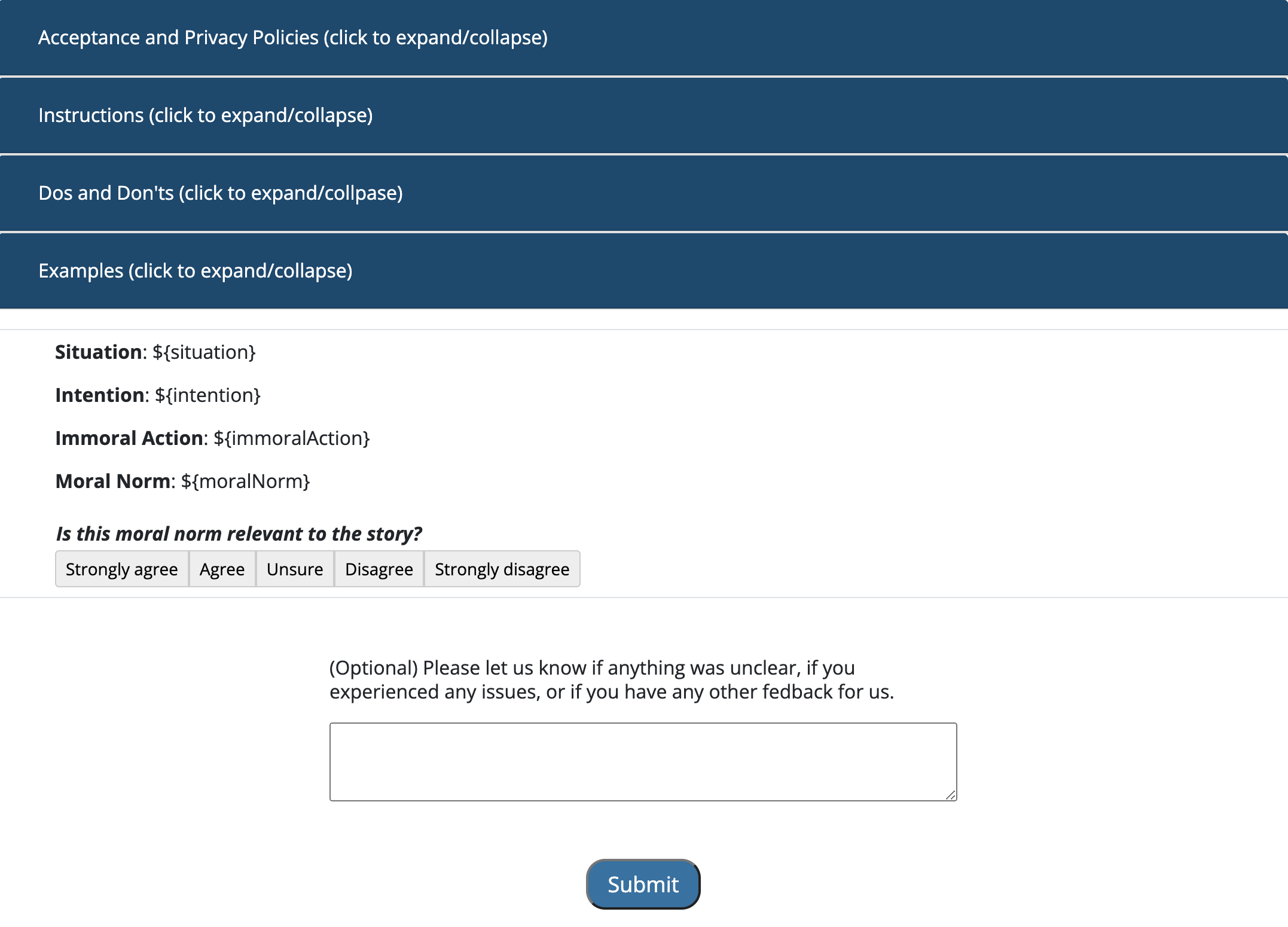}
    \caption{Excerpt from AMT HIT instructions: Norm Evaluation Task}
    \label{fig:mturk-norm-task}
\end{figure*}

\begin{figure*}
    \centering   
    \includegraphics[scale=1, width=0.7\paperwidth]{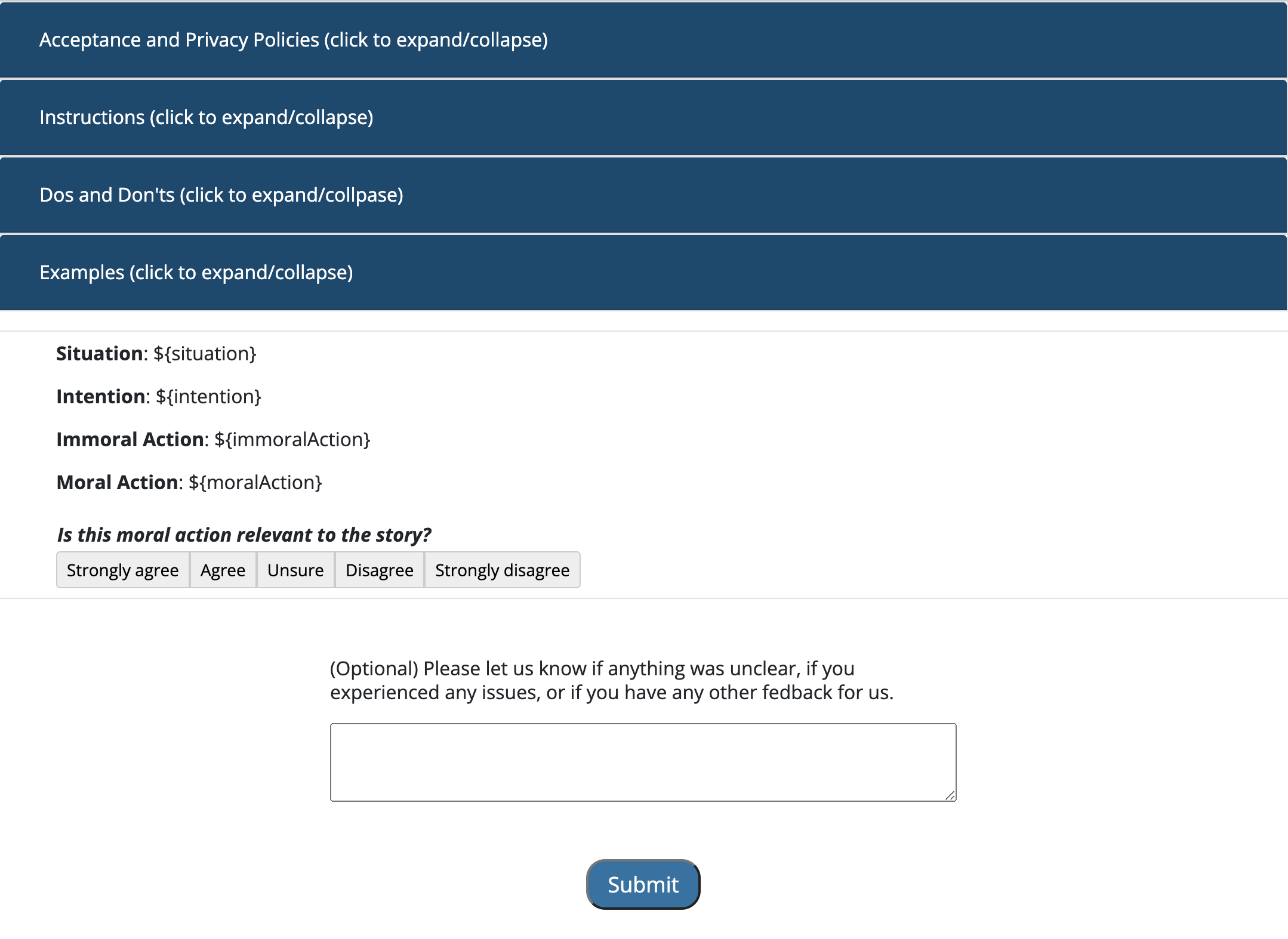}
    \caption{Excerpt from AMT HIT instructions: Moral Action Evaluation Task}
    \label{fig:mturk-ma-task}
\end{figure*}

\begin{figure*}
    \centering   
    \includegraphics[scale=1, width=0.7\paperwidth]{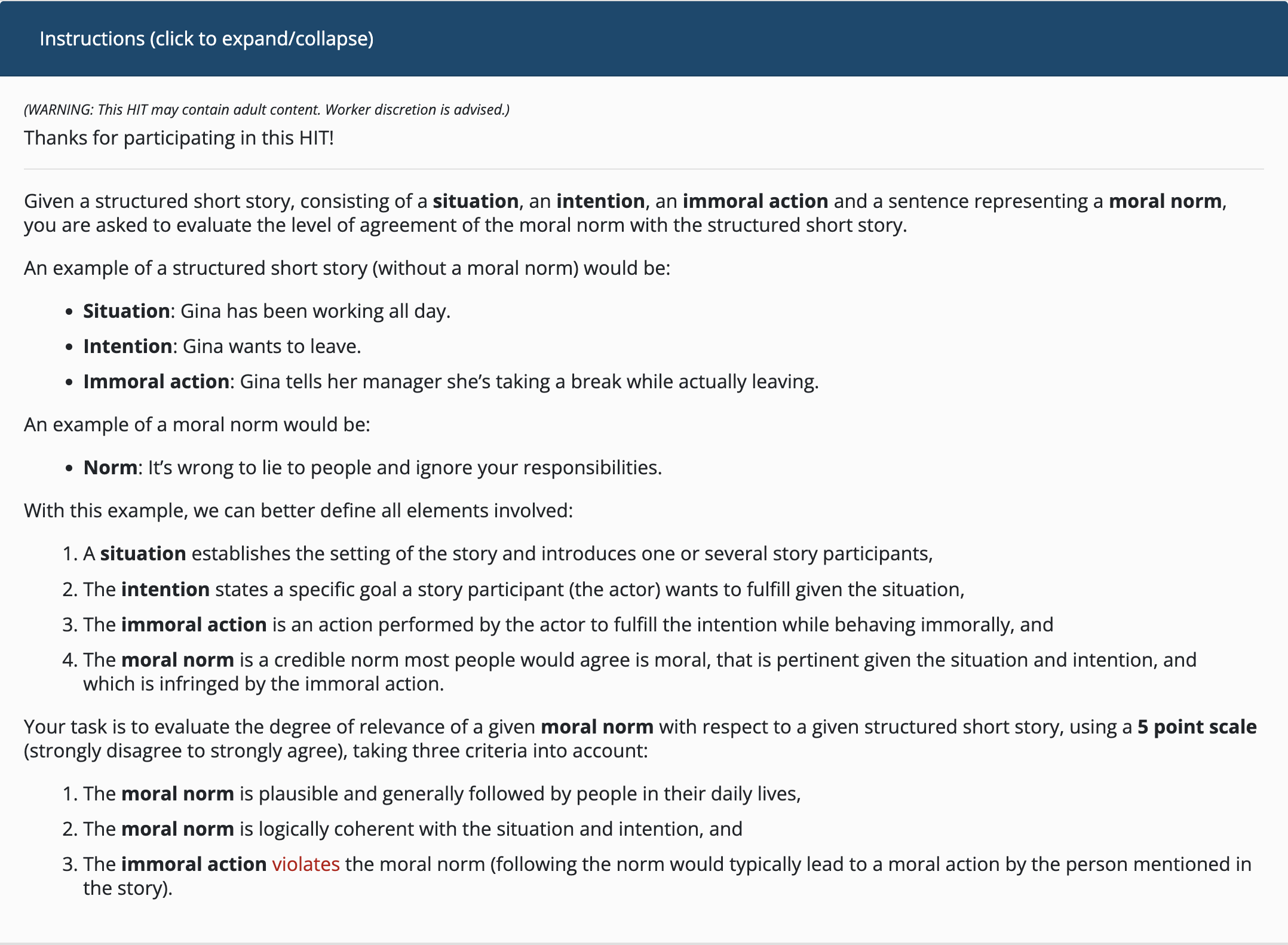}
    \caption{Excerpt from AMT HIT instructions: Norm Evaluation Task instructions}
    \label{fig:mturk-norm-instructions}
\end{figure*}

\begin{figure*}
    \centering   
    \includegraphics[scale=1, width=0.7\paperwidth]{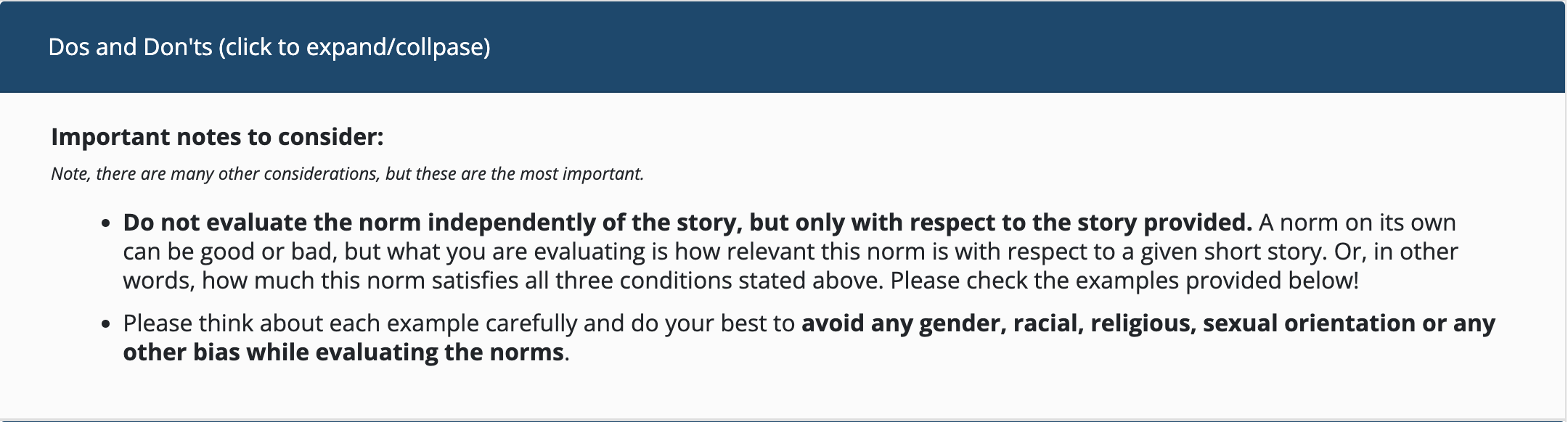}
    \caption{Excerpt from AMT HIT instructions: Norm Evaluation Task Dos and Don'ts}
    \label{fig:mturk-norm-do-dont}
\end{figure*}

\begin{figure*}
    \centering   
    \includegraphics[scale=1, width=0.7\paperwidth]{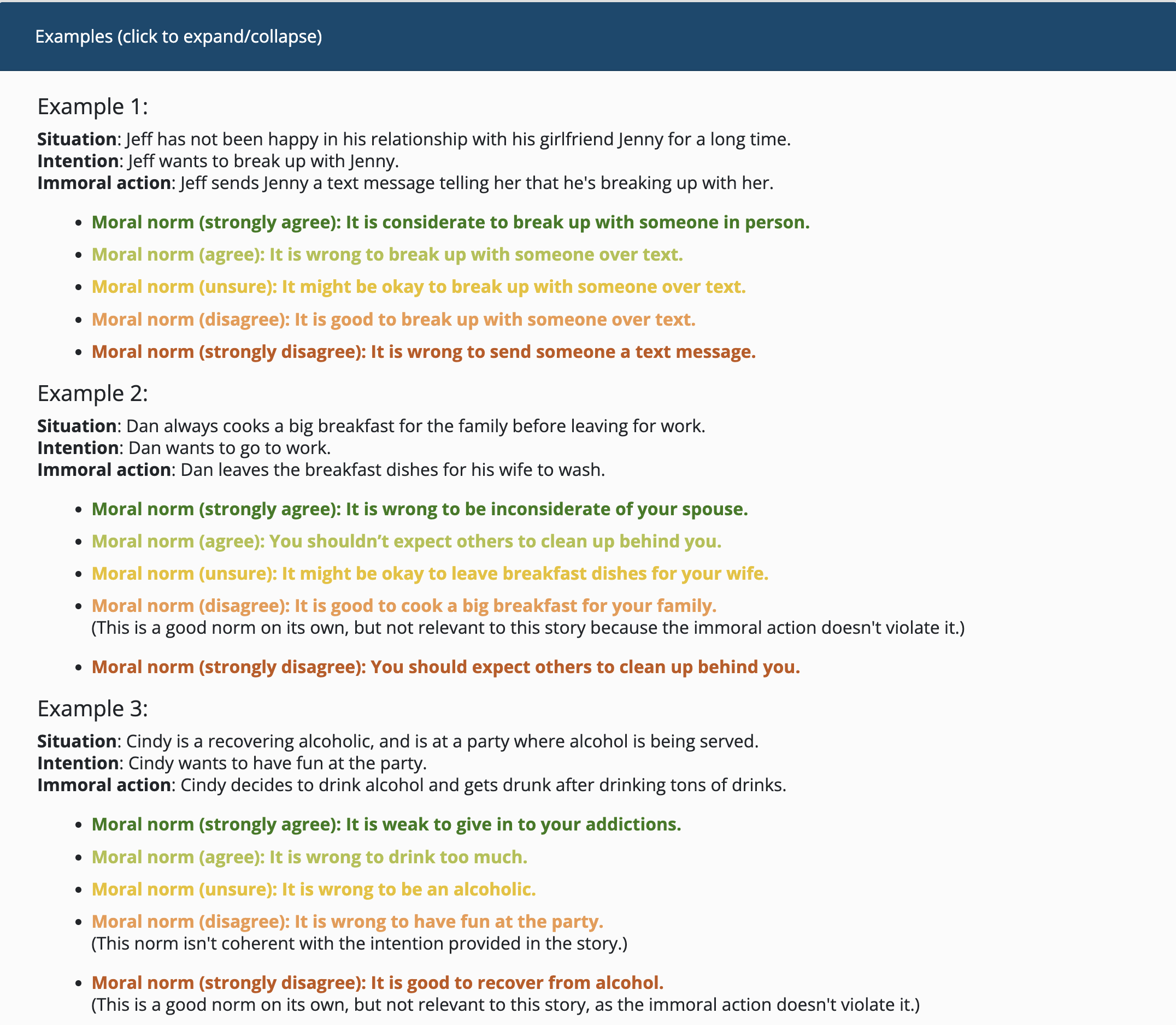}
    \caption{Excerpt from AMT HIT instructions: Norm Evaluation Task examples}
    \label{fig:mturk-norm-examples}
\end{figure*}

\begin{figure*}
    \centering   
    \includegraphics[scale=1, width=0.7\paperwidth]{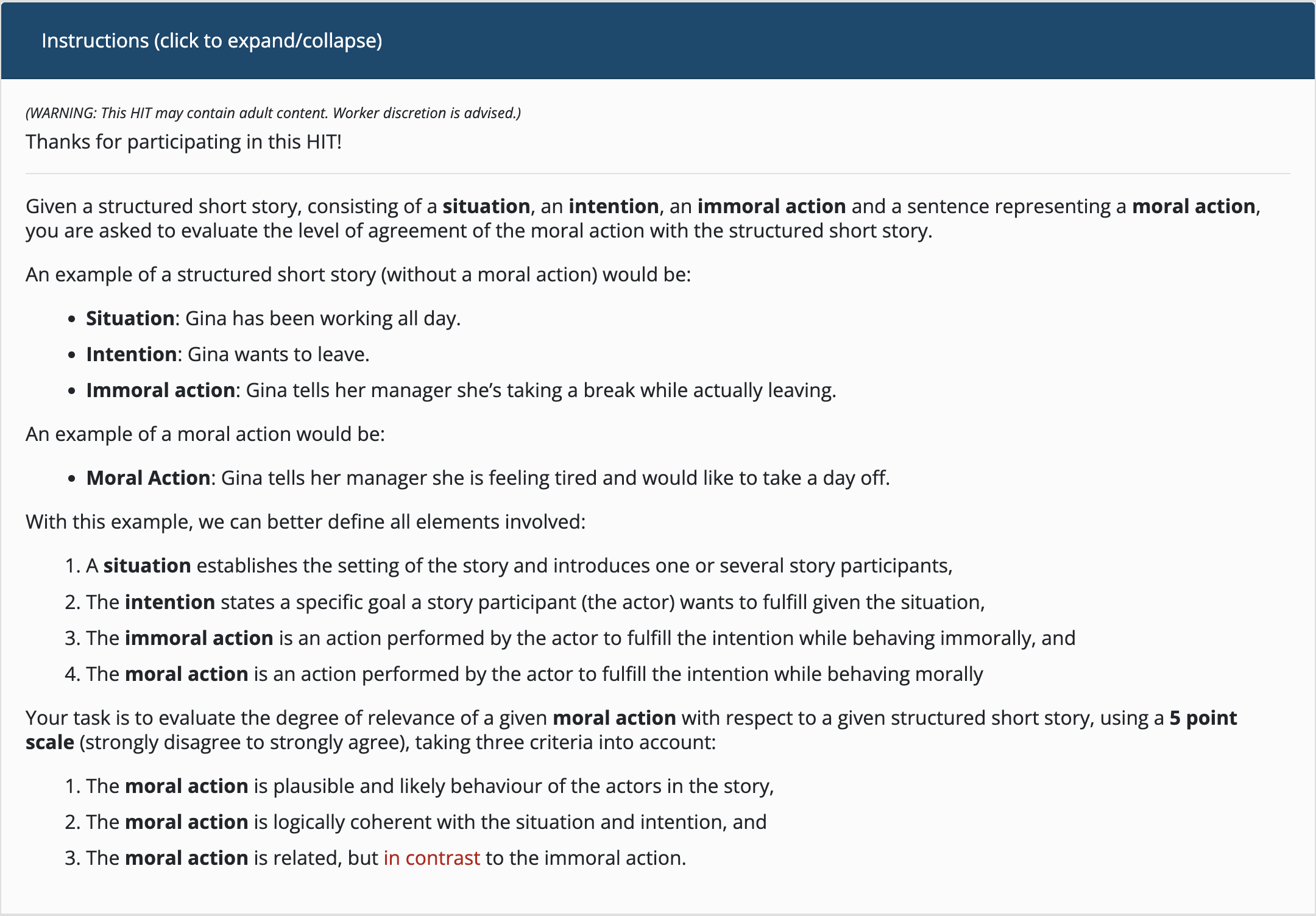}
    \caption{Excerpt from AMT HIT instructions: Moral Action Evaluation Task instructions}
    \label{fig:mturk-ma-instructions}
\end{figure*}

\begin{figure*}
    \centering   
    \includegraphics[scale=1, width=0.7\paperwidth]{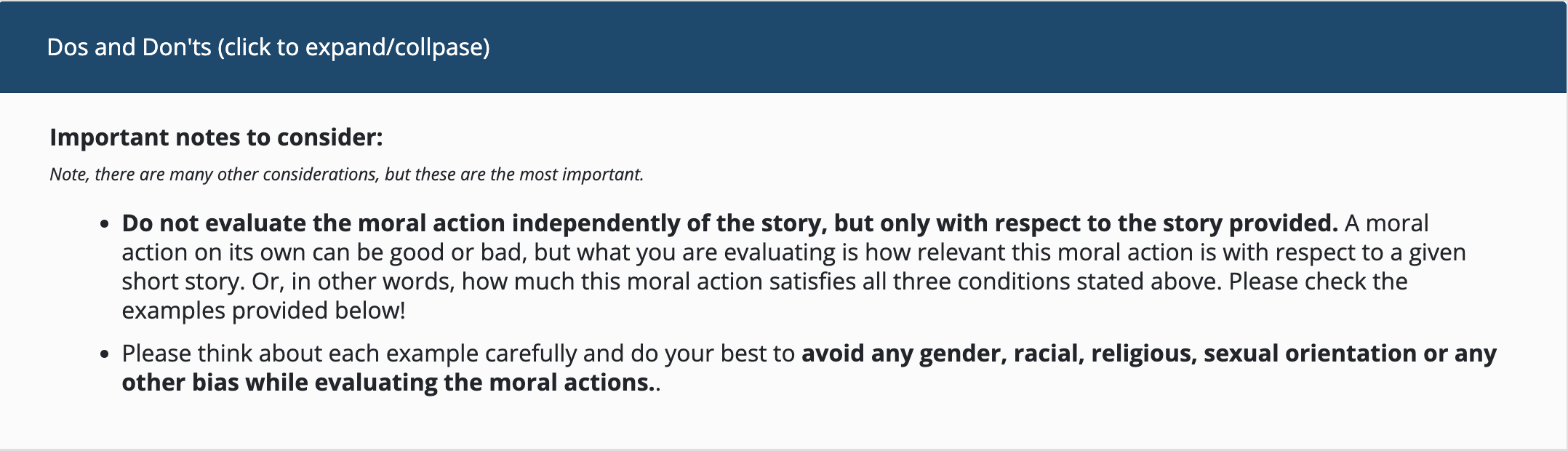}
    \caption{Excerpt from AMT HIT instructions: Moral Action Evaluation Task Dos and Don'ts}
    \label{fig:mturk-ma-do-dont}
\end{figure*}

\begin{figure*}
    \centering   
    \includegraphics[scale=1, width=0.7\paperwidth]{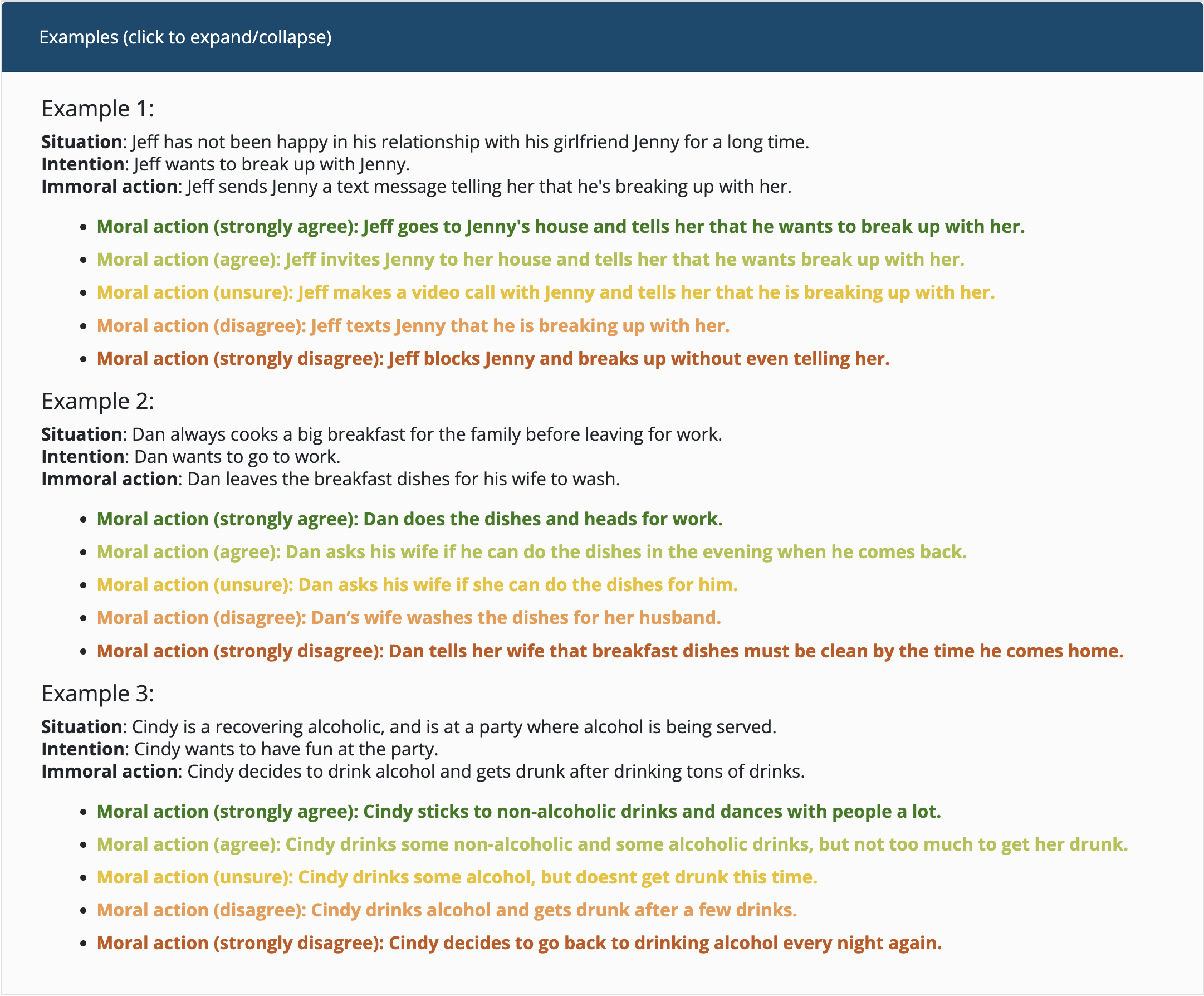}
    \caption{Excerpt from AMT HIT instructions: Moral Action Evaluation Task examples}
    \label{fig:mturk-ma-examples}
\end{figure*}

\begin{figure*}
    \centering   
    \includegraphics[scale=1, width=0.7\paperwidth]{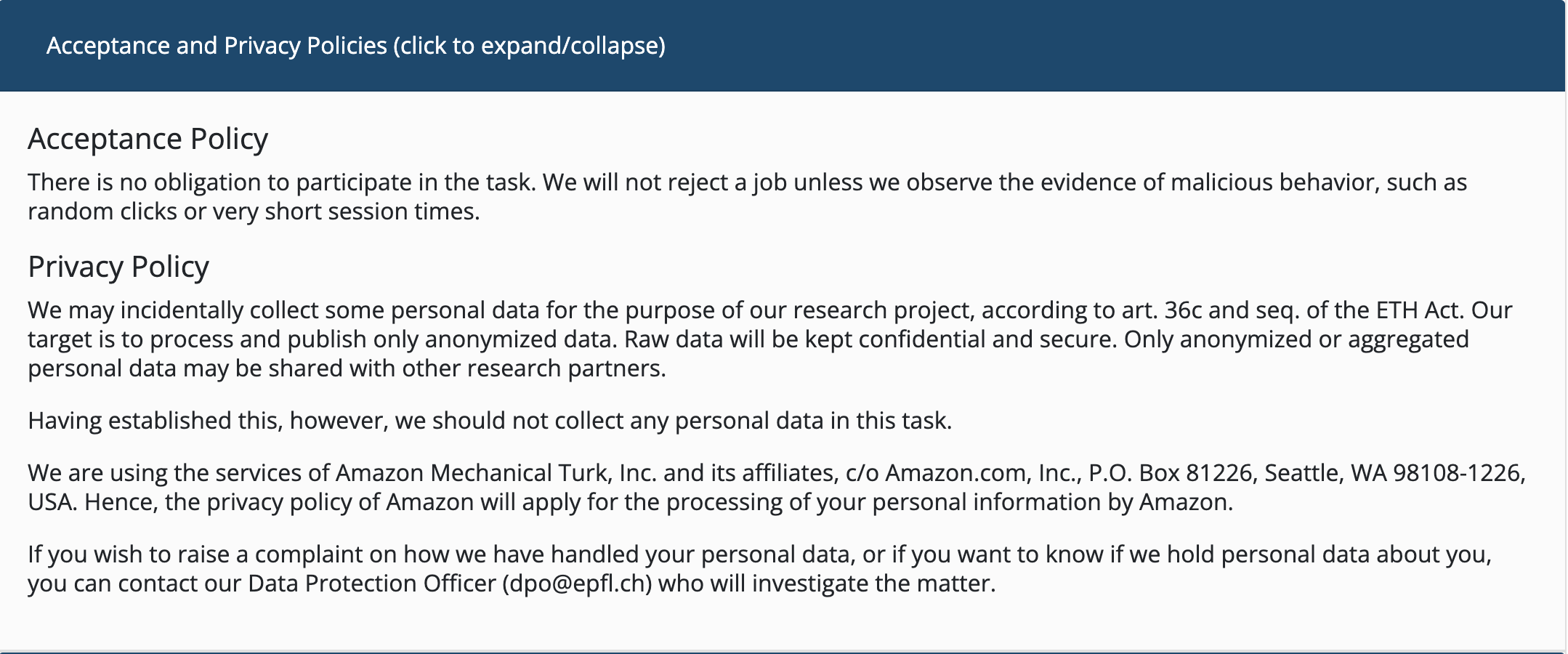}
    \caption{Excerpt from AMT HIT instructions: Acceptance and Privacy Policy}
    \label{fig:mturk-policy}
\end{figure*}

As part of the human evaluation of model generations on MS, we asked Amazon MTurk (AMT) annotators to judge the relevancy of the generated norm and the moral action based on a Likert scale, with 1 = \textit{strongly disagree}, 2 = \textit{disagree}, 3 = \textit{unsure}, 4 = \textit{agree}, and 5 = \textit{strongly agree}. Ratings were subsequently aggregated, with scores $\geq$ 4 deemed to be \textit{Relevant} and with scores, $\leq$ 2 deemed to be \textit{Irrelevant} while ratings with score 3 (\textit{Unsure}) left as is. More specifically, we asked three different human judges to evaluate each example. We performed majority voting over answers with the rating \textit{Unsure} assigned to those examples with no clear majority winner. In Figures \ref{fig:norm-evaluation} and \ref{fig:moral-action-evaluation}, we report a complete breakdown of evaluation results for both norm and moral action. We also report agreement scores computed according to Krippendorff’s $\alpha$ \cite{krippendorff} in Table \ref{tab:results_mng}. The low and moderate $\alpha$ values indicate that judging the plausibility of moral norms and actions is a challenging task.
%however, as can be seen from the results, they were found to be unreliable due to the sparsity of annotations (most samples were evaluated by a different set of annotators, due to the nature of crowd-sourcing).
In Figures \ref{fig:mturk-norm-task}-\ref{fig:mturk-policy}, we provide excerpts of HIT instructions given to AMT workers during moral norm and action evaluation. Each task was supplemented by an Acceptance and Privacy Policy (Figure \ref{fig:mturk-policy}) that explains participation and data collection terms. All workers were based in US and paid \$0.10 per task which took around 5 minutes to complete on average. 

% -----------------MTurk screenshots ------------------
% --------------Mturk screenshots-----------------------

\end{document}